\documentclass{article}

\usepackage[final, nonatbib]{neurips_2019}
\let\citep\cite

\usepackage{authblk}

\usepackage{booktabs}       
\usepackage{amsfonts}       
\usepackage{microtype}      
\usepackage{xcolor}
\usepackage{url}
\usepackage{verbatim} 
\usepackage{graphicx}
\usepackage{caption} 
\usepackage{multirow}
\usepackage[linesnumbered,ruled]{algorithm2e}
\usepackage{xspace}
\usepackage{epsfig}
\usepackage{amsmath}
\usepackage{amsthm}
\usepackage{amssymb}
\usepackage{xr}
\usepackage{bbm}
\usepackage{bm}
\usepackage{subcaption}
\usepackage{enumitem}

\usepackage{enumitem}
\setlist[enumerate]{noitemsep, topsep=0.5\topsep}
\setlist[description]{noitemsep, topsep=0.5\topsep}
\setlist[itemize]{noitemsep, topsep=0.5\topsep}

\newcommand{\jure}[1]{{{\textcolor{red}{[Jure: #1]}}}}
\newcommand{\marinka}[1]{{{\textcolor{purple}{[M: #1]}}}}
\newcommand{\rex}[1]{{{\textcolor{magenta}{[Rex: #1]}}}}

\newcommand{\dylan}[1]{{{\textcolor{green}{[Dylan: #1]}}}}
\newcommand{\hide}[1]{}

\newcommand{\CITE}{{\textcolor{red}{[CITE]}}}
\newcommand{\name}{\textsc{GnnExplainer}\xspace}
\newcommand{\namelong}{\textsc{GnnExplainer}\xspace}
\newcommand{\longname}{\textsc{GnnExplainer}\xspace}
\newcommand{\gnn}{\textrm{GNN}\xspace}
\newcommand{\gnns}{\textrm{GNNs}\xspace}
\newcommand{\lexp}{\textrm{SI-Exp}\xspace}
\newcommand{\gexp}{\textrm{MI-Exp}\xspace}

\newcommand{\cut}[1]{}
\newcommand{\xhdr}[1]{\vspace{-1mm}\noindent{{\bf #1.}}}

\newcommand{\eg}{\emph{e.g.}\xspace}
\newcommand{\ie}{\emph{i.e.}\xspace}

\DeclareMathOperator*{\argmax}{argmax}

\usepackage{pifont}

\newcommand{\ba}{Barab\'asi-Albert\xspace}

\title{GNNExplainer: Generating Explanations \\for Graph Neural Networks}

\begin{document}

 \author[$\dagger$]{Rex~Ying}
 \author[$\dagger$,$\ddagger$]{Dylan~Bourgeois}
 \author[$\dagger$]{Jiaxuan~You}
 \author[$\dagger$]{Marinka~Zitnik}
 \author[$\dagger$]{Jure~Leskovec}
 \affil[$\dagger$]{Department of Computer Science, Stanford University}
 \affil[$\ddagger$]{Robust.AI}
 \affil[ ]{\footnotesize{\texttt{\{rexying, dtsbourg, jiaxuan, marinka, jure\}@cs.stanford.edu}}}

\maketitle

%
\begin{abstract}
Graph Neural Networks (GNNs) are a powerful tool for machine learning on graphs.
GNNs combine node feature information with the graph structure by 
recursively passing neural messages along edges of the input graph. However, incorporating
both graph structure and feature information leads to complex models
and explaining predictions made by GNNs remains unsolved. Here
we propose {\em \namelong}, the first general, model-agnostic approach for providing
interpretable explanations for predictions of any GNN-based model on any graph-based machine learning task. 
Given an instance, {\em \namelong} identifies a compact subgraph structure and a small subset of node features that have a crucial role in GNN's prediction. 
Further, {\em \namelong} can generate consistent and concise explanations for an entire class of instances.
We formulate {\em \namelong} as an optimization task that maximizes the mutual information between a GNN's prediction and distribution of possible subgraph structures. 
Experiments on synthetic and real-world graphs show that our approach can identify important graph structures as well as node features, and outperforms alternative baseline approaches by up to 43.0\% in explanation accuracy. \namelong provides a variety of benefits, from the ability to visualize semantically relevant structures to interpretability, to giving insights into errors of faulty GNNs.
\end{abstract}

\hide{
Graph Neural Networks (GNNs) are a powerful tool for machine learning on graphs.
GNNs combine node feature information with the graph structure by using neural
networks to pass messages through edges of the graph.  However, incorporating
both graph structure and feature information leads to complex non-linear models,
and explaining predictions made by GNNs remains an unsolved problem. Here
we propose {\em \namelong}, the first general, model-agnostic approach for providing
interpretable explanations for predictions of any GNN-based model on any graph-
based machine learning task.  In order to explain a given node’s predicted label,
{\em \namelong} provides {\em single-instance explanations} by highlighting relevant
features as well as an important subgraph structure by identifying the edges that
are most relevant to the prediction. Additionally, the model provides {\em multi-instance
explanations} that aim to explain predictions for an entire class of instances/nodes.
We formalize {\em \namelong} as an optimization task that maximizes the mutual
information between the prediction of the full model and the distribution of subgraph structure explanation. We experiment on synthetic as well as real-world data, showing that our approach is able to highlight relevant topological structures. \namelong provides a variety of benefits, from the identification of semantically relevant structures to explaining predictions, to providing guidance when debugging faulty graph neural network models.
}

\cut{
Graph Neural Networks (GNNs) are a powerful tool for machine learning on graphs. GNNs combine node feature information with the graph structure by using neural networks to pass messages through edges in the graph.
However, incorporating both graph structure and feature information leads to complex non-linear models and explaining predictions made by GNNs remains to be a challenging task.
Here we propose GnnExplainer, 
a general model-agnostic approach for providing interpretable explanations for predictions of any GNN-based model on any graph-based machine learning task (node and graph classification, link prediction).
In order to explain a given node's predicted label, GnnExplainer provides a local interpretation by highlighting relevant features as well as an important subgraph structure by identifying the edges that are most relevant to the prediction. Additionally, the model provides single-instance explanations when  given a single prediction as well as multi-instance explanations that aim to explain predictions for an entire class of instances/nodes.
We formalize GnnExplainer as an optimization task that maximizes the mutual information between the prediction of the full model and the prediction of simplified explainer model.
We experiment on synthetic as well as real-world data. On synthetic data we demonstrate that our approach is able to highlight relevant topological structures from noisy graphs. We also demonstrate GnnExplainer to provide a better understanding of pre-trained models on real-world tasks. GnnExplainer provides a variety of benefits, from the identification of semantically relevant structures to explain predictions to providing guidance when debugging faulty graph neural network models.
}

\cut{
Graph Neural Networks (GNNs) are a powerful tool for machine learning on graphs. GNNs combine node feature information with the graph structure by using neural networks to pass messages through edges in the graph.
However, incorporating both graph structure and feature information leads to complex non-linear models and explaining predictions made by GNNs remains to be a challenging task.
Here we propose {\em \namelong}, 
a general model-agnostic approach for providing interpretable explanations for predictions of any \gnn-based model on any graph-based machine learning task.
In order to explain a given node's predicted label, \namelong provides a {\em single-instance explanations} by highlighting relevant features as well as an important subgraph structure by identifying the edges that are most relevant to the prediction. Additionally, the model provides {\em multi-instance explanations} that aim to explain predictions for an entire class of instances/nodes.
We formalize \name as an optimization task that maximizes the mutual information between the prediction of the full model and the prediction of simplified explainer model.
We experiment on synthetic as well as real-world data, showing that our approach is able to highlight relevant topological structures from noisy graphs. 
\name provides a variety of benefits, from the identification of semantically relevant structures to explain predictions to providing guidance when debugging faulty graph neural network models.}



\section{Introduction}

\hide{
\begin{figure}[h]
    \centering
    \includegraphics[width=0.6\columnwidth]{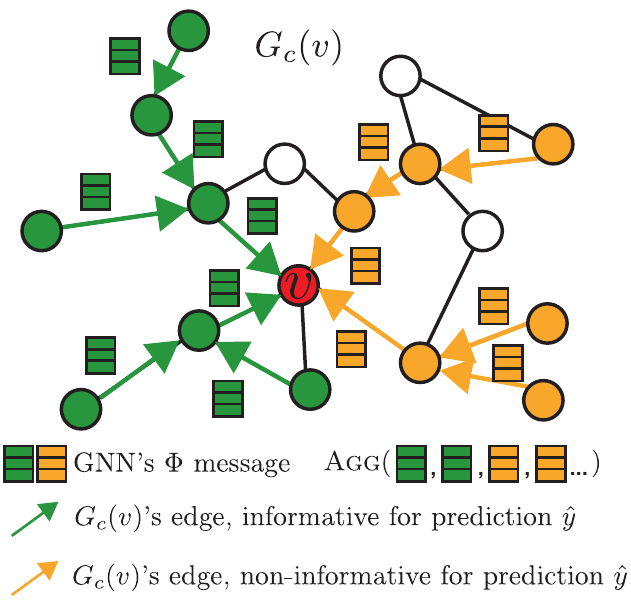}
    \vspace{-2mm}
    \caption{\gnn computation graph $G_c$ for making a prediction $\hat{y}$ at node $v$. Some edges form important message-passing pathways (green) while others do not (orange). \gnn model aggregates informative as well as non-informative messages to form a prediction at node $v$. The goal of \name is to identify a small set of important features and pathways (green) that are crucial for prediction.
    }
    \label{fig:explainer-motivation}
    \vspace{-4mm}
\end{figure}
}

In many real-world applications, including social, information, chemical, and biological domains, data can be naturally modeled as graphs~\citep{cho2011friendship,you2018graph,zitnik2018decagon}. Graphs are powerful data representations but are challenging to work with because they require modeling of rich relational  information as well as node feature information~\citep{zhang_deep_2018,zhou_graph_2018}.
%
%
To address this challenge, Graph Neural Networks (\gnns) have emerged as state-of-the-art for machine learning on graphs, due to their ability to recursively incorporate information from neighboring nodes in the graph, naturally capturing both graph structure and node features~\citep{graphsage,kipf2016semi,ying2018hierarchical,zhang2018link}.

Despite their strengths, {\gnn}s lack transparency as they do not easily allow for a human-intelligible explanation of their predictions.
Yet, the ability to understand \gnn's predictions is important and useful for several reasons: (i) it can increase trust in the \gnn model, (ii) it improves model's transparency in a growing number of decision-critical applications pertaining to fairness, privacy and other safety challenges~\citep{doshi-velez_towards_2017}, and (iii) it allows practitioners to get an understanding of the network characteristics, identify and correct systematic patterns of mistakes made by models before deploying them in the real world.



\begin{figure}[t]
    \centering
    \includegraphics[width=\columnwidth]{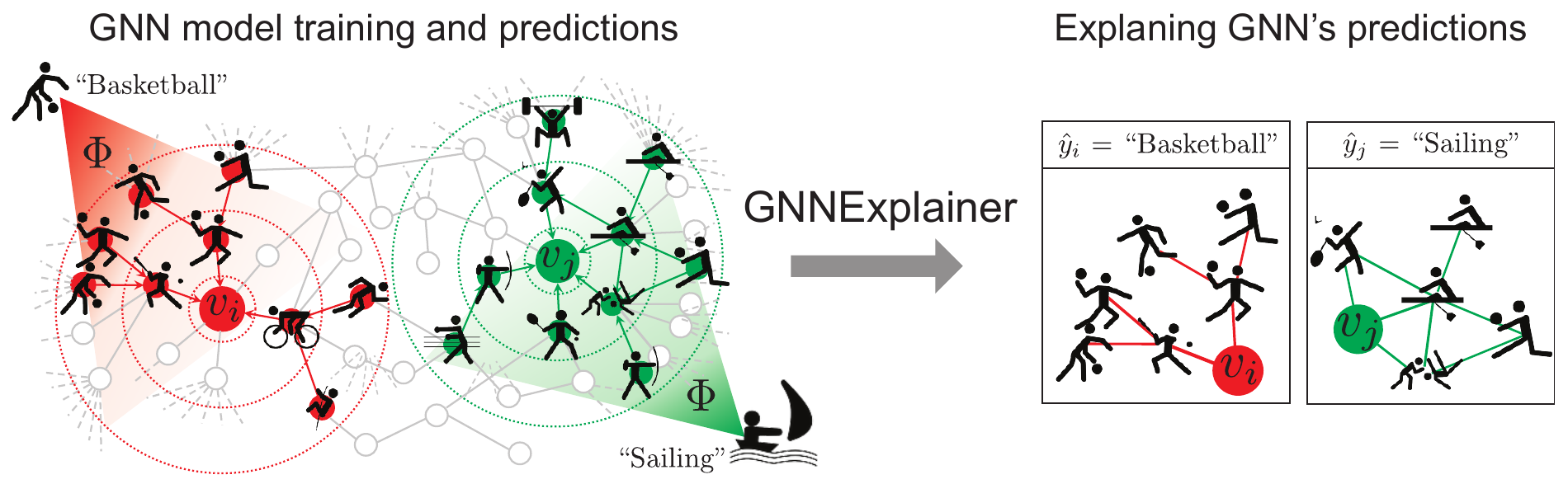}
    \vspace{-6mm}
    \caption{\name provides interpretable explanations for predictions made by any GNN model on any graph-based machine learning task. Shown is a hypothetical node classification task where a GNN model $\Phi$ is trained on a social interaction graph to predict future sport activities. Given a trained GNN $\Phi$ and a prediction $\hat{y}_i$ = ``Basketball'' for person $v_i$, \name generates an explanation by identifying a small subgraph of the input graph together with a small subset of node features (shown on the right) that are most influential for $\hat{y}_i$. Examining explanation for $\hat{y}_i$, we see that many friends in one part of $v_i$'s social circle enjoy ball games, and so the GNN predicts that $v_i$ will like basketball. Similarly, examining explanation for $\hat{y}_j$, we see that $v_j$'s friends and friends of his friends enjoy water and beach sports, and so the GNN predicts $\hat{y}_j$ = ``Sailing.''
 }
    \label{fig:explainer-intro}
    \vspace{-6mm}
\end{figure}

While currently there are no methods for explaining \gnns, recent approaches for explaining other types of neural networks have taken one of two main routes. One line of work locally approximates models with simpler surrogate models, which are then probed for explanations~\citep{lakkaraju_interpretable_2017,ribeiro_why_2016, schmitz_ann-dt:_1999}. Other methods carefully examine models for relevant features and find good qualitative interpretations of high level features~\citep{chen2018learning,Erhan2009VisualizingHF,lundberg_unified_2017, sundararajan_axiomatic_nodate} or identify influential input instances~\citep{koh_understanding_2017, DBLP:journals/corr/abs-1811-09720}.
However, these approaches fall short in their ability to incorporate relational information, the essence of graphs. Since this aspect is crucial for the success of machine learning on graphs, any explanation of {\gnn}'s predictions\textbf{} should leverage rich relational information provided by the graph as well as node features.

Here we propose {\em \namelong}, an approach for explaining predictions made by \gnns. \name takes a trained \gnn and its prediction(s), and it returns an explanation in the form of a small subgraph of the input graph together with a small subset of node features that are most influential for the prediction(s) (Figure~\ref{fig:explainer-intro}). The approach is model-agnostic and can explain predictions of any \gnn on any machine learning task for graphs, including node classification, link prediction, and graph classification.  
It handles single- as well as multi-instance explanations. In the case of single-instance explanations, \name explains a \gnn's prediction for one particular instance (\ie, a node label, a new link, a graph-level label). In the case of multi-instance explanations, \name provides an explanation that consistently explains a set of instances (\eg, nodes of a given class). 


\name specifies an explanation as a rich subgraph of the entire graph the \gnn was trained on, such that the subgraph maximizes the mutual information with \gnn's prediction(s). This is achieved by formulating a mean field variational approximation and learning a real-valued {\em graph mask} which selects the important subgraph of the \gnn's computation graph. Simultaneously, \name also learns a {\em feature mask} that masks out unimportant node features (Figure~\ref{fig:explainer-intro}). 



\cut{
\jure{The issue with this paragraph was that it was too focused on numbers (quantitative comparison). But given that we are the first method to do explanatations, I think the focuse should be claiming it works well and only cite numbers as evidence. I fixed it as follows:}}
We evaluate \namelong on synthetic as well as real-world graphs. 
Experiments show that \name provides consistent and concise explanations of \gnn's predictions. 
On synthetic graphs with planted network motifs, which play a role in determining node labels, we show that \name accurately identifies the subgraphs/motifs as well as node features that determine node labels outperforming alternative baseline approaches by up to 43.0\% in explanation accuracy. Further, using two real-world datasets we show how \name can provide important domain insights by robustly identifying important graph structures and node features that influence a \gnn's predictions. Specifically, using molecular graphs and social interaction networks, we show that \name can identify important domain-specific graph structures, such as $NO_2$ chemical groups or ring structures in molecules, and star structures in Reddit threads.
Overall, experiments demonstrate that \name provides consistent and concise explanations for \gnn-based models for different machine learning tasks on graphs.

\hide{
We extensively validate \namelong on synthetic and real-world graphs. Experiments show that \name provides consistent and concise explanations of \gnn's predictions. We demonstrate our method on both single- as well as multi-instance explanations. Using carefully designed synthetic data with ``planted'' pathways important for prediction, we show that \name can accurately identify important topological structures used by the given \gnn. Furthermore, we show \name can also robustly identify most important features that influence \gnn's prediction the most. We also demonstrate \name on two real-world datasets: molecule classification and social network classification. We show that \name is able to explain the graph structures that a given \gnn model learned to use for prediction. For example, in the molecule classification task \name identified important and domain-relevant topological structures, such as $NO_2$ functional groups and ring structures in molecules. Overall, experiments demonstrate that \name provides consistent and concise explanations for \gnn-based models for different machine learning tasks on graphs.
}

\hide{
In handling these data structures, initial approaches relied on handcrafted features and measures. Following a similar trend as the machine learning community at large, manual heuristics have progressively been replaced with data-driven, automatic feature extractors. These networks learn to decide what is relevant in a given prediction task thanks to learnable filters, which can incorporate node features but also the inductive bias provided by the edges connecting instances.

This differentiable formulation, known as Graph Neural Networks ({\gnn}s), is able to handle a wide range of tasks, across domains. They are able to learn global graph representations, which can be leveraged to classify entire graphs, for example to predict the anti-cancer properties of a given molecule. One of the most popular settings however is node classification, wherein a label is to be inferred for individual nodes on the graph. Applications of this task are abundant, for example predicting the customer type in an online purchase graph or the role of a protein in a biological network. 

Many early successes came in problems with a semi-supervised formulation, where the graph is fixed but there are missing node labels within it. The power of deep learning on graphs has been further demonstrated by the development of new methods that work in the inductive setting, where the classifier is able to generalize to entirely unseen graphs. To do so, the network must be able to effectively learn not only from the node's feature but also incorporate information from its neighbours, leveraging the graph topology to do so. However, obvious successes in terms of task accuracy have been tarnished by a lack of transparency into the results: neural models remain notoriously difficult to interpret. As deep learning on graphs progresses to more critical applications, \eg, drug-discovery or medical diagnoses, this interpretability component becomes crucial to ensure trust in the model's predictions.
}



\section{Related work}
\label{sec:related}


Although the problem of explaining GNNs is not well-studied, the related problems of interpretability and neural debugging received substantial attention in machine learning. At a high level, we can group those interpretability methods for non-graph neural networks into two main families.  

Methods in the first family formulate simple proxy models of full neural networks. This can be done in a model-agnostic way, usually by learning a locally faithful approximation around the prediction, for example through linear models~\citep{ribeiro_why_2016} or sets of rules, representing sufficient conditions on the prediction~\citep{augasta_reverse_2012,lakkaraju_interpretable_2017,calders_deepred_2016}. 
Methods in the second family identify important aspects of the computation, for example, through feature gradients~\citep{Erhan2009VisualizingHF,fleet_visualizing_2014}, backpropagation of neurons' contributions to the input features~\citep{chen2018learning,shrikumar_learning_2017,sundararajan_axiomatic_nodate}, and counterfactual reasoning~\citep{Kang2019explaine}. 
However, the saliency maps~\citep{fleet_visualizing_2014} produced by these methods have been shown to be misleading in some instances~\citep{2018sanity} and prone to issues like gradient saturation~\citep{shrikumar_learning_2017,sundararajan_axiomatic_nodate}. These issues are exacerbated on discrete inputs such as graph adjacency matrices since the gradient values can be very large but only on very small intervals. Because of that, such approaches are not suitable for explaining predictions made by neural networks on graphs.

Instead of creating new, inherently interpretable models, post-hoc interpretability methods~\citep{adadi_peeking_2018,fisher_all_2018,guidotti_survey_2018,hooker_discovering_2004,koh_understanding_2017,DBLP:journals/corr/abs-1811-09720} consider models as black boxes and then probe them for relevant information. 
However, no work has been done to leverage relational structures like graphs. The lack of methods for explaining predictions on graph-structured data is problematic, as in many cases, predictions on graphs are induced by a complex combination of nodes and paths of edges between them. For example, in some tasks, an edge is important only when another alternative path exists in the graph to form a cycle, and those two features, only when considered together, can accurately predict node labels ~\citep{mutag,duvenaud_convolutional_2015}. Their joint contribution thus cannot be modeled as a simple linear combinations of individual contributions. 

Finally, recent \gnn models augment interpretability via attention mechanisms~\citep{neil2018interpretable, velickovic2018graph,PhysRevLett.120.145301}. However, although the learned edge attention values can indicate important graph structure, the values are the same for predictions across all nodes. Thus, this contradicts with many applications where an edge is essential for predicting the label of one node but not the label of another node. Furthermore, these approaches are either limited to specific \gnn architectures or cannot explain predictions by jointly considering both graph structure and node feature information.

\hide{
Methods in the first family formulate a simpler proxy model for the full neural network. This can be done in a model-agnostic way, usually by learning a locally faithful approximation around the prediction, for example with a linear model~\cite{ribeiro_why_2016} or a set of rules, representing sufficient conditions on the prediction~\cite{augasta_reverse_2012,calders_deepred_2016,lakkaraju_interpretable_2017}. 
Global distillations of the main model have also been proposed, for instance by reducing deep neural networks to decision trees~\cite{calders_deepred_2016, schmitz_ann-dt:_1999}. However, such approaches often produce intractably large surrogate models, which in practice are uninterpretable.

A second family of models instead aims to highlight relevant aspects of the computation within the provided model. The main approach here is to inspect feature gradients~\cite{Erhan2009VisualizingHF} but many other related ideas have also been proposed~\cite{sundararajan_axiomatic_nodate, shrikumar_learning_2017}. When overlayed to the input data, these methods produce a saliency map~\cite{fleet_visualizing_2014} which reveals important features or raw pixels. However, saliency maps have been shown to be misleading in some instances~\cite{2018sanity} and prone to issues such as gradient saturation~\cite{sundararajan_axiomatic_nodate, shrikumar_learning_2017}. These issues are exacerbated on discrete inputs such as graph adjacency matrices, since the gradient values can be very large but on a very small interval. This means such approaches are unsuitable for explaining relational structure of a \gnn, which is our goal here.

Last, algorithms that find patterns of the input data~\cite{koh_understanding_2017, DBLP:journals/corr/abs-1811-09720} to identify influential samples are an example of post-hoc interpretability methods. Instead of creating new, inherently interpretable models, thse approaches consider the model as a black box~\cite{guidotti_survey_2018, adadi_peeking_2018} and then probe it for relevant information. Most techniques isolate individual input samples, with some methods allowing for important interactions to be highlighted~\cite{fisher_all_2018, hooker_discovering_2004}. However, no work has been done to leverage stronger relational structures like graphs. In contrast, in many cases prediction on graphs can be induced by a complex composition of nodes and their paths. For example, in some tasks an edge could be important only when another alternative path exists to form a cycle, which determines the class of the node. Therefore their joint contribution cannot be modeled well using linear combinations of individual contributions. 
}

\hide{
Recent instances of \gnn models have been proposed to augment the interpretability properties of \gnns~\cite{PhysRevLett.120.145301, neil2018interpretable, velickovic2018graph}. However, in contrast to our work here, \rex{these approaches are limited to their own \gnn architecture, and cannot make explanations using both features and subgraph structure jointly.}
\rex{Notably, although the edge attention values in graph attention networks can serve as indication of structure importance, it is the same for predictions on all nodes. Thus it contradicts with many application scenario where an edge is important for predictions on one node, but not the other.
(or maybe move to experiments)}
\rex{remove?} Our goal here is different, as we are already given a trained model in the \gnn-family and aim to explain its predictions.
}

\hide{
\xhdr{Graph Neural Networks} Graph Neural Networks (\gnn)~\cite{scarselli} obtain node embeddings by recursively propagating information from its neighbours. This framework was later unified into a general Neural Message-Passing scheme~\cite{gilmer2017neural}, and more recently into the relational inductive bias model~\cite{battaglia}. For a more detailed review of recent developments we please refer the reader to~\cite{zhang_deep_2018, zhou_graph_2018, battaglia,hamilton2017representation}.
Under this model, \gnns have achieved state-of-the-art performance across a variety of tasks, such as node classification~\cite{kipf2016semi, graphsage}, link prediction~\cite{zhang2018link, schlichtkrull2018modeling}, graph clustering~\cite{defferrard2016convolutional, ying2018hierarchical} or graph classification~\cite{ying2018hierarchical, dai2016discriminative, duvenaud_convolutional_2015}. These tasks occur in domains where the graph structure is ubiquitous, such as social networks~\cite{backstrom2011supervised}, content graphs~\cite{pinsage}, biology~\cite{agrawal2018large}, and chemoinformatics~\cite{duvenaud_convolutional_2015,jin2017predicting,zitnik2018decagon}.
Recent instances of \gnn models have been proposed to augment the interpretability properties of \gnns~\cite{PhysRevLett.120.145301, neil2018interpretable, velickovic2018graph}. However, in contrast to our work here, \rex{these approaches are limited to their own \gnn architecture, and cannot make explanations using both features and subgraph structure jointly.}
\rex{Notably, although the edge attention values in graph attention networks can serve as indication of structure importance, it is the same for predictions on all nodes. Thus it contradicts with many application scenario where an edge is important for predictions on one node, but not the other.
(or maybe move to experiments)}
\rex{remove?} Our goal here is different, as we are already given a trained model in the \gnn-family and aim to explain its predictions.

}

\section{Formulating explanations for graph neural networks}
\label{sec:explainer}

\begin{figure*}[t]
    \centering
    \includegraphics[width=0.9\textwidth]{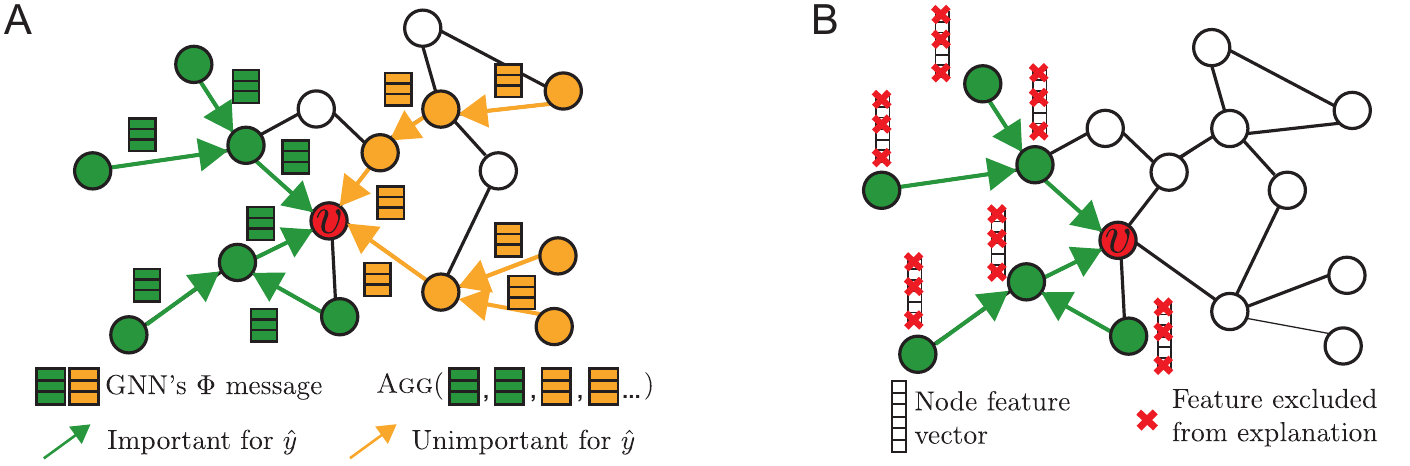}
    \vspace{-2mm}
    \caption{\textbf{A.} \gnn computation graph $G_c$ (green and orange) for making prediction $\hat{y}$ at node $v$. Some edges in $G_c$ form important neural message-passing pathways (green), which allow useful node information to be propagated across $G_c$ and aggregated at $v$ for prediction, while other edges do not (orange). However, \gnn needs to aggregate important as well as unimportant messages  to form a prediction at node $v$, which can dilute the signal accumulated from $v$'s neighborhood.  The goal of \name is to identify a small set of important features and pathways (green) that are crucial for prediction. \textbf{B.} In addition to $G_S$ (green), \name identifies what feature dimensions of $G_S$'s nodes are important for prediction
    by learning a node feature mask.}
    \label{fig:definition-node-features}
    \vspace{-5mm}
\end{figure*}

Let $G$ denote a graph on edges $E$ and nodes $V$ that are associated with $d$-dimensional node features $\mathcal{X} = \{x_1, \ldots, x_n\}$, $x_i \in \mathbb{R}^d$. Without loss of generality, we consider the problem of explaining a node classification task (see Section~\ref{sec:other-tasks} for other tasks). 
Let $f$ denote a label function on nodes $f: V \mapsto \{1, \ldots, C\}$ that maps every node in $V$ to one of $C$ classes. 
The \gnn model $\Phi$ is optimized on all nodes in the training set and is then used for prediction, \ie, to approximate $f$ on new nodes.

\subsection{Background on graph neural networks}
\label{sec:background}

At layer $l$, the update of \gnn model $\Phi$ involves three key computations~\citep{battaglia,zhang_deep_2018,zhou_graph_2018}. (1) First, the model computes neural messages between every pair of nodes. The message for node pair $(v_i, v_j)$ is a function \textsc{Msg} of $v_i$'s  and $v_j$'s representations $\mathbf{h}_i^{l-1}$ and $\mathbf{h}_j^{l-1}$ in the previous layer and of the relation $r_{ij}$ between the nodes: $m_{ij}^l = \textsc{Msg}(\mathbf{h}_i^{l-1}, \mathbf{h}_j^{l-1}, r_{ij}).$
(2)
Second, for each node $v_i$, \gnn aggregates messages from $v_i$'s neighborhood $\mathcal{N}_{v_i}$ and calculates an aggregated message $M_i$ via an aggregation method \textsc{Agg}~\citep{graphsage,xu2018powerful}: 
$M_{i}^l = \textsc{Agg}(\{m_{ij}^l | v_j \in \mathcal{N}_{v_i}\}),$
where $\mathcal{N}_{v_i}$ is neighborhood of node $v_i$ whose definition depends on a particular GNN variant. 
(3) Finally, \gnn takes the aggregated message $M_i^l$ along with $v_i$'s representation $\mathbf{h}_i^{l-1}$ from the previous layer, and it non-linearly transforms them to obtain $v_i$'s representation $\mathbf{h}_i^l$ at layer $l$:
$\mathbf{h}_{i}^l  = \textsc{Update}(M_i^l, \mathbf{h}_i^{l-1}).$
%
%
%
The final embedding for node $v_i$ after $L$ layers of computation is $\mathbf{z}_i = \mathbf{h}_i^L$. 
Our \name provides explanations for any \gnn that can be formulated in terms of \textsc{Msg}, \textsc{Agg}, and \textsc{Update} computations. 



\hide{
%
In the context of a neural network architecture, the information that a \gnn relies on for computing $\mathbf{z}_i$ is completely determined by $v_i$'s computation graph, which is defined by Eq.~\ref{eq:agg}. The \gnn uses that computation graph to generate $v_i$'s representation $\mathbf{z}_i$ (Eqs.~\ref{eq:msg} and~\ref{eq:update}). Importantly, the structure of the computation graph is different for each node $v_i$, and depends on how the neighborhood $\mathcal{N}_{v_i}$ is defined. Let $G_c(v_i)$ denote the computation graph used by the \gnn to compute representation $\mathbf{z}_i$ of node $v_i$.
The $G_c(v_i)$ can be obtained by performing a graph traversal of arbitrary depth $L$, \eg, a Breadth-First Search (BFS), using $\mathcal{N}_{v_i}$ as the neighborhood definition. We further define $A_c(v_i)$ as the adjacency matrix corresponding to the computation graph $G_c(v_i)$.

Following this definition, in Graph Convolutional Networks (GCNs)~\citep{kipf2016semi}, the computation graph $G_c(v_i)$ is simply an $L$-hop neighborhood of $v_i$ in the input graph $G$. However, in other \gnn models, such as Jumping Knowledge Networks~\citep{xujumping}, attention-based networks~\cite{velickovic2018graph}, and Line-Graph NNs~\citep{chen2018supervised}, the computation graph $G_c(v_i)$ will be different from the exhaustive $L$-hop neighborhood.
}

\hide{
\subsection{Desired Features of \name}
\label{subsec:desiderata}


An essential criterion for explanations is that they must be interpretable, \ie, provide a qualitative understanding of the relationship between the input nodes and the prediction. Such a requirement implies that explanations should be easy to understand while remaining exhaustive. This means that a \gnn explainer should take into account both the structure of the underlying graph $G$ as well as the associated features, if they are available. More specifically, the following aspects of a \gnn model should be incorporated into the design of explanations:
\begin{enumerate}[leftmargin=*]
\item \textbf{Local edge fidelity:}
The explanation needs to identify the relational inductive bias used by the \gnn. This means it needs to identify which message-passing edges in the computation graph $G_c$ represent essential relationships for a prediction.
\item \textbf{Local node fidelity:}
The explanation should not only incorporate the specific node $v_i$'s features $\mathbf{x}_i$, but also a number of important features from the set $X_c(v_i)$ of features from other nodes present in computation graph $G_c(v_i)$.
\item \textbf{Single-instance vs. multi-instance explanations:}
The explanation should summarize where in a graph $G$ the \gnn model looks for evidence for its prediction and identify the subgraph of $G$ most responsible for a given prediction. The \gnn explainer should also be able to provide an explanation for a set of predictions, \eg, by capturing  a distribution of computation graphs for all nodes that belong to the same class.
\item \textbf{Any \gnn model:} 
A \gnn explainer should be able to explain {\em any} model in \gnn-family and be model-agostic, \ie, treat \gnn as a black box without requiring modifications of neural architecture or re-training.
\item \textbf{Any prediction task on graphs:}
A \gnn explainer should be applicable to {\em any} machine learning task on graphs: node classification, link prediction, and graph classification. \rex{these are repeating the intro}
\end{enumerate}
%
}

\subsection{\name: Problem formulation}


%
Our key insight is the observation that the computation graph of node $v$, which is defined by the GNN's neighborhood-based aggregation (Figure~\ref{fig:definition-node-features}), fully determines all the information the \gnn uses to generate prediction $\hat{y}$ at node $v$. In particular, $v$'s computation graph tells the GNN how to generate $v$'s embedding $\mathbf{z}$. Let us denote that computation graph by $G_c(v)$, the associated binary adjacency matrix by $A_c(v) \in \{0,1\}^{n \times n}$, and the associated feature set by $X_c(v) = \{x_j | v_j \in G_c(v)\}$. The \gnn model $\Phi$ learns a conditional distribution $P_\Phi(Y | G_c, X_c)$, where $Y$ is a random variable representing labels $\{1, \ldots, C\}$, indicating the probability of nodes belonging to each of $C$ classes. 

A \gnn's prediction is given by $\hat{y} = \Phi(G_c(v), X_c(v))$, meaning that it is fully determined by the model $\Phi$, graph structural information $G_c(v)$, and node feature information $X_c(v)$. In effect, this observation implies that we only need to consider graph structure $G_c(v)$ and node features $X_c(v)$ to explain $\hat{y}$ (Figure~\ref{fig:definition-node-features}A).
Formally, \name generates explanation for prediction $\hat{y}$ as $(G_S,X_S^F)$, where $G_S$ is a small subgraph of the computation graph. $X_S$ is the associated feature of $G_S$, and $X_S^F$ is a small subset of node features (masked out by the mask $F$, \ie, $X_S^F = \{x_j^F|v_j \in G_S\}$) that are most important for explaining $\hat{y}$ (Figure~\ref{fig:definition-node-features}B).

\hide{
The information that a \gnn~ relies on for computing $\mathbf{z}_i$ can be completely described by its computation graph $G_c(v_i)$ and the associated feature set $X_c(v_i) = \{x_j | v_j \in G_c(v_i)\}.$
%
We thus use these elements as inputs to the \gnn~model $\hat{y} = \Phi(G_c(v_i), X_c(v_i))$.
Therefore, when constructing a local explanation for the prediction a \gnn has made on $v_i$, we only consider structural information present in $G_c(v_i)$, and node features from $X_c(v_i)$.
}


\hide{
\subsection{Problem formulation}

\marinka{
Moved from background, it is crucial for defining single-instance explanations:
The information that a \gnn~ relies on for computing $\mathbf{z}_i$ can be completely described by its computation graph $G_c(v_i)$ and the associated feature set $X_c(v_i) = \{x_j | v_j \in G_c(v_i)\}.$
%
We thus use these elements as inputs to the \gnn~model $\hat{y} = \Phi(G_c(v_i), X_c(v_i))$.
Therefore, when constructing a local explanation for the prediction a \gnn has made on $v_i$, we only consider structural information present in $G_c(v_i)$, and node features from $X_c(v_i)$.
}

In this section, we concretely describe the setting in which \name is used for explaining \gnn~ predictions.
Without loss of generality, we use \name to explain the \gnn~ predictions of a node classification task, but \name can be easily adapted to the task of link prediction and graph classification.

In a node classification task, a graph, defined as $G = (V, E)$, a set of nodes $V$ connected through edges $E$, is associated with a set of node features $\mathcal{X} = \{\mathbf{x}_1, \ldots, \mathbf{x}_n\}$, $x_i \in \mathbb{R}^d$, and a label function on nodes $f: V \mapsto \{1, \ldots, C\}$ that maps every node in $V$ to one of the $C$ classes. 
The \gnn~ model $\Phi$ is optimized for all nodes in the training set, such that $\Phi$ can be used to approximate $f$ on nodes in the test set. During this phase model is only exposed to $G=(V, E_{\mathrm{train}})$, leaving a subset of edges and labels on nodes unobserved.

Once the \gnn~ model $\Phi$ is trained, we can harness \name as a tool to analyze individual node predictions, but also as the decision boundary of node classes, in order to provide insights on what the model has learned. Given the architecture of GNN models, this explanation will be supported by highlighting which edge connectivity patterns and node features are leveraged the most in the \gnn~ predictions.

In our setting, we assume that the \gnn~ model $\Phi$ is already trained, and that \name is able to access $\Phi$ in terms of forward and backward propagation. This places \name in the family of post-hoc interpretability methods.

\name relies on this general formulation, but is agnostic to the specific form of \gnn . 
We then discuss the desired form of explanations for \gnn s, which present notable differences from the explanations of other classes of models.

\subsection{Overview of GNN-Explainer}

In this work, we introduce GNN-Explainer, a tool for post-hoc interpretation of predictions generated by a pre-trained \gnn. The approach is able to identify nodes that are locally relevant to a given prediction, similarly to existing interpretation methods. However, it is also able to identify relevant structures in the induced computation graph, which is a sample of a given node's neighborhood. We also present an extension to provide insight into global structures in the graph that are relevant to a given class of nodes. 

To achieve these goals, we define a useful \gnn~ explanation as being composed of two key components: a local explainer (\lexp) and a global explainer (\gexp). The local explainer \lexp~ identifies, among the computation graph $G_c(v_i)$ of each node $v_i$, the particular feature dimensions and message-passing pathways that most contribute to the prediction made by the model for the given node's label.

In other words, we aim to identify a subgraph of the computation graph $G_c(v_i)$, among all possible subgraphs of $G_c(v_i)$, and a subset of feature dimensions for node features, that is the most influential in the model's prediction.
In this process both important message-passing and node features are jointly determined, due to the message aggregation scheme of \gnn s, which incorporates both aspects in a single forward pass.

In a second step, the global explainer \gexp~ will aggregate, in a class-specific way, information from the local explainer \lexp. This way, the local explanations are summarized into a prototype computation graph. The global explainer \gexp~ therefore allows the comparison of different computation graphs across different nodes in graphs, offering insights into the distribution of computation graphs for nodes in a certain label class.

\rex{How to use explainer}
}
\section{\name}

\label{sec:methods}

\hide{
We start by describing single-instance explanations that consider graph structure and node feature information (Sections~\ref{sec:local-model} and \ref{sec:incorporate-features}) and proceed with multi-instance explanations (Section~\ref{sec:global-model}). We conclude with a discussion on how \name can be used for any machine learning task on graphs including link prediction and graph classification (Section~\ref{sec:other-tasks}) and we give details on \name training (Section~\ref{sec:training-efficiency}).
}

Next we describe our approach {\em \namelong}.
Given a trained \gnn model $\Phi$ and a prediction (\ie, single-instance explanation, Sections~\ref{sec:local-model} and \ref{sec:incorporate-features}) or a set of predictions (\ie, multi-instance explanations, Section~\ref{sec:global-model}), the \name will generate an explanation by identifying a subgraph of the computation graph and a subset of node features that are most influential for the model $\Phi$'s prediction. 
In the case of explaining a set of predictions, \name will aggregate individual explanations in the set and automatically summarize it with a prototype. We conclude this section with a discussion on how \name can be used for any machine learning task on graphs, including link prediction and graph classification (Section~\ref{sec:other-tasks}).


\hide{
Our \name is a post-hoc method for explaining \gnn predictions. Given a trained \gnn model $\Phi$ and a prediction (\ie, single-instance explanation) or a set of predictions (\ie, multi-instance explanations), the \name will generate an explanation by identifying a subgraph of the computation graph $G_c(v_i)$ and a subset of node features that are most influential for the model $\Phi$'s prediction. 
In the case of explaining a set of predictions, \name will aggregate individual explanations in the set and summarize them into a prototype computation graph.
}

\hide{
Next we describe our approach {\em \namelong} that satisfies desiderata 1--5 from Section~\ref{subsec:desiderata}. We start by describing single-instance explanations (Section~\ref{sec:local-model}) that consider structural graph patterns and rich node features (Section~\ref{sec:incorporate-features}) and proceed with multi-instance explanations based on the idea of class prototypes (Section~\ref{sec:global-model}). We conclude with a discussion on how \name can be used for any machine learning task on graphs including link prediction and graph classification (Section~\ref{sec:other-tasks}) and we give details on \name training (Section~\ref{sec:training-efficiency}).
}

\hide{
\xhdr{Overview of \longname}
\name is a tool for post-hoc interpretation of predictions generated by a pre-trained \gnn. It is formulated in an optimization framework. The key insight here is that information used by \gnn to compute prediction $\hat{y}$ is completely described by its computation graph $G_c(v_i)$ and the associated feature set $X_c(v_i) = \{x_j | v_j \in G_c(v_i)\}$ (Section~\ref{sec:background}); that is, \gnn prediction is given as $\hat{y} = \Phi(G_c(v_i), X_c(v_i))$. As a result, to construct an explanation for $\hat{y}$ we only need to consider structural information present in $G_c(v_i)$ and node feature information present in $X_c(v_i)$.




\name identifies a subgraph of the computation graph $G_c(v_i)$ and a subset of node features that are most influential for the model's prediction. Importantly, influential edges and node features in $G_c(v_i)$ are jointly determined. In the case of a set of predictions, \name aggregates, in a class-specific way, individual explanations in the set. This way, individual explanations are summarized into a prototype computation graph. 



}

\subsection{Single-instance explanations}
\label{sec:local-model}

Given a node $v$, our goal is to identify a subgraph $G_S \subseteq G_c$ and the associated features $X_S = \{x_j | v_j \in G_S\}$ that are important for the \gnn's prediction $\hat{y}$. For now, we assume that $X_S$ is a small subset of $d$-dimensional node features; we will later discuss how to automatically determine which dimensions of node features need to be included in explanations (Section~\ref{sec:incorporate-features}). We formalize the notion of importance using mutual information $MI$ and formulate the \name as the following optimization framework:
\begin{equation}
    \label{eq:mi_obj}
    \max_{G_S } MI\left(Y, (G_S, X_S)\right) = H(Y) - H(Y | G=G_S, X=X_S).
\end{equation}
\hide{
\rex{rationale of MI? 
The underlying assumption is that the prediction made by the model after softmax coorelates with the probability of belonging to each class. Although it's for single instance, the model predicts the probability based on all nodes during training.
MI has been widely used to evaluate the correlation between random variables. 
}
}
For node $v$, $MI$ quantifies the change in the probability of prediction $\hat{y} = \Phi(G_c, X_c)$ when $v$'s computation graph is limited to explanation subgraph $G_S$ and its node features are limited to $X_S$. 

For example, consider the situation where $v_j \in G_c(v_i),$ $v_j \neq v_i$. Then, if removing $v_j$ from $G_c(v_i)$ strongly decreases the probability of prediction $\hat{y}_i$, the node $v_j$ is a good counterfactual explanation for the prediction at $v_i$. Similarly, consider the situation where $(v_j, v_k) \in G_c(v_i),$ $v_j, v_k \neq v_i$. Then, if removing an edge between $v_j$ and $v_k$ strongly decreases the probability of prediction $\hat{y}_i$ then the absence of that edge is a good counterfactual explanation for the prediction at $v_i$. 

Examining Eq.~(\ref{eq:mi_obj}), we see that the entropy term $H(Y)$ is constant because $\Phi$ is fixed for a trained GNN. As a result, maximizing mutual information between the predicted label distribution $Y$ and explanation $(G_S, X_S)$ is equivalent to minimizing conditional entropy $H(Y | G=G_S, X=X_S)$, which can be expressed as follows:
\begin{equation}
    \label{eq:ent_obj}
    \!\!\!H(Y | G\!=\!G_S, X\!=\!X_S) = -\mathbb{E}_{Y | G_S, X_S} \left[ \log P_\Phi (Y | G\!=\!G_S, X\!=\!X_S) \right].
\end{equation} 
Explanation for prediction $\hat{y}$ is thus a subgraph $G_S$ that minimizes uncertainty of $\Phi$ when the GNN computation is limited to $G_S$. In effect, $G_S$ maximizes probability of $\hat{y}$ (Figure~\ref{fig:definition-node-features}). To obtain a compact explanation, we impose a constraint on $G_S$'s size as: $|G_S| \le K_M,$ so that $G_S$ has at most $K_M$ nodes. In effect, this implies that \name aims to denoise $G_c$ by taking $K_M$ edges that give the highest mutual information with the prediction. 
%

\hide{
Intuitively, one can motivate the process of generating explanations as follows. Some edges in $v_i$'s computation graph $G_c$ form important message-passing pathways, which allow useful node information to be propagated across $G_c$ and aggregated at $v_i$ for prediction. However, some edges in $G_c$ might not be informative for prediction. In particular, note that the \gnn is trained on a number of examples/nodes and our aim is to explain the prediction of a particular node $v_i$. Thus, not all parts of the \gnn $\Phi$ might be relevant or important for prediction at $v_i$. Hence messages passed along those edges do not provide any useful information and might even decrease \gnn's performance. Indeed, the aggregator in Equation~\ref{eq:agg} needs to aggregate such non-informative messages together with useful messages, which can dilute the overall signal accumulated from $v_i$'s neighborhood. In practice, many datasets and tasks support such motivation as illustrated in Figure~\ref{fig:explainer-motivation}. The objective of \name thus aims to remove edges from $v_i$'s computation graph that are not informative for the $v_i$'s prediction. In other words, the objective is in a sense denoising the computation graph to keep at most $k$ edges that have the highest mutual information between the computation graph and the \gnn's prediction. 
}



\hide{
\xhdr{Connection to Feature Selection}
The motivation for \lexp is also similar to feature selection explanation~\cite{Erhan2009VisualizingHF, chen2018learning, sundararajan_axiomatic_nodate, lundberg_unified_2017}, whose aim is to select a subset of feature dimensions that are most informative to the model prediction. However, an important distinction is that the importance of feature dimensions can usually be treated as independent: the importance of one dimension does not depend on the importance of the other.\dylan{The dependency is across nodes, not really feature dimensions, right?}
In contrast, the importance of edges in \gnn~ subgraphs are correlated.
A simple example is that: an edge could be important when there are another alternative pathway that forms a cycle, indicating a special structural role of the node to be predicted. Hence the decision to include this edge in $G_S$ depends on whether the alternate pathway is included. \rex{do we need figure to illustrate?}

Furthermore, in the feature selection setting, the feature dimension of different data points is fixed.
However, since the computation graph $G_c(v_i)$ is different for explanation of different node $v_i$, the choice of $G_S$ for each node $v_i$ does not have an exact correspondence. 
As a consequence, the optimization of the \lexp objective has to be local and specific to the individual computation graph $G_c(v_i)$.
} 

\xhdr{\name's optimization framework}
Direct optimization of \name's objective is not tractable as $G_c$ has exponentially many subgraphs $G_S$ that are candidate explanations for $\hat{y}$.
We thus consider a fractional adjacency matrix\footnote{For typed edges, we define $G_S \in [0,1]^{C_e \times n \times n}$ where $C_e$ is the number of edge types. 
} for subgraphs $G_S$, \ie, $A_S \in [0,1]^{n\times n}$, and enforce the subgraph constraint as: $A_S[j,k] \le A_c[j,k]$ for all $j, k$.
%
This continuous relaxation can be interpreted as a variational approximation of distribution of subgraphs of $G_c$. In particular, if we treat $G_S \sim \mathcal{G}$ as a random graph variable, the objective in Eq.~(\ref{eq:ent_obj}) becomes:
\begin{equation}
    \label{eq:variational_relaxation}
    \min_{\mathcal{G}} \mathbb{E}_{G_S \sim \mathcal{G}} H(Y | G=G_S, X=X_S),
\end{equation}
%
With convexity assumption, Jensen's inequality gives the following upper bound:
\begin{equation}
    \label{eq:variational_upper_bound}
    \min_{\mathcal{G}} H(Y | G=\mathbb{E}_{\mathcal{G}}[G_S], X=X_S).
\end{equation}
In practice, due to the complexity of neural networks, the convexity assumption does not hold. However, experimentally, we found that minimizing this objective with regularization often leads to a local minimum corresponding to high-quality explanations.

To tractably estimate $\mathbb{E}_{\mathcal{G}}$, we use mean-field variational approximation and decompose $\mathcal{G}$ into a multivariate Bernoulli distribution as: $P_\mathcal{G}(G_S) = \prod_{(j,k) \in G_c} {A_S}[j,k]$. This allows us to estimate the expectation with respect to the mean-field approximation, thereby obtaining $A_S$ in which $(j, k)$-th entry represents the expectation on whether edge $(v_j, v_k)$ exists.
We observed empirically that this approximation together with a regularizer for promoting discreteness~\citep{ying2018hierarchical} converges to good local minima despite the non-convexity of \gnns.
%
%
The conditional entropy in Equation \ref{eq:variational_upper_bound} can be optimized by replacing the $\mathbb{E}_{\mathcal{G}}[G_S]$ to be optimized by a masking of the computation graph of adjacency matrix, $A_c \odot \sigma(M)$, where $M \in \mathbb{R}^{n\times n}$ denotes the mask that we need to learn, $\odot$ denotes element-wise multiplication, and $\sigma$ denotes the sigmoid that maps the mask to $[0,1]^{n\times n}$.

In some applications, instead of finding an explanation in terms of model's confidence, the users care more about ``why does the trained model predict a certain class label'', or ``how to make the trained model predict a desired class label''. We can modify the conditional entropy objective in Equation \ref{eq:variational_upper_bound} with a cross entropy objective between the label class and the model prediction\footnote{The label class is the predicted label class by the GNN model to be explained, when answering ``why does the trained model predict a certain class label''. ``how to make the trained model predict a desired class label'' can be answered by using the ground-truth label class.}.
To answer these queries, a computationally efficient version of \name's objective, which we optimize using gradient descent, is as follows:
\begin{equation}
    \min_{M}  -\sum_{c =1}^C \mathbbm{1}[y = c] \log P_\Phi (Y = y | G=A_c \odot \sigma(M), X=X_c), \label{eq:opt_node_class}
\end{equation}
The masking approach is also found in Neural Relational Inference~\cite{kipf2018neural}, albeit with different motivation and objective.
Lastly, we compute the element-wise multiplication of $\sigma(M)$ and $A_c$ and remove low values in $M$ through thresholding to arrive at the explanation $G_S$ for the GNN model's prediction $\hat{y}$ at node $v$.
%
%
\hide{
\marinka{The following is unclear}
In the forward propagation, $P_\Phi (Y | G=A_S \odot M, X=X_S)$ is computed by replacing adjacency matrix $A_c$ with $A_S \odot M$ in the aggregation step \ref{eq:agg}, meaning that the mask controls how much information is passed along each edge of $G_c$. An alternative to continuous relaxation is Gumbel Softmax, which allows learning of discrete adjacency $A_S$. We empirically compare both approaches in Section~\ref{sec:exp}.
}
%


\subsection{Joint learning of graph structural and node feature information}
\label{sec:incorporate-features}


\hide{
Node feature information plays an important role in computing messages between nodes. The explanations therefore should take into account the feature information. However, in a \gnn node feature information is propagated over multiple layers, which means important features could differ between the nodes and the features might interact in non-linear ways across the multi-layer \gnn. We tackle this problem by learning a feature mask (one for the entire \gnn or one per node in the \gnn computation graph). The mask then acts as a feature selector and also prevents the feature information from ``leaking'' through the layers of the \gnn.
}


\hide{
We explicitly consider what feature dimensions are essential for message passing, \emph{i.e.} feature selection for nodes in $G_S$. 
In addition to optimizing Eq.~(\ref{eq:ent_obj}) over $G_S$, we also perform optimization that selects important features in $X_S$. 
Two approaches are possible: (1) we can optimize the essential dimensions for each node's feature $x_i \in G_S$ separately, or (2) we can optimize a single set of important dimensions for all nodes' features involved in the \gnn computation graph.
The former allows for finer granularity of feature importance but provides more complex explanations, while the latter is computationally more efficient and provides more concise explanations. Here we adopt the latter approach, but the following analysis can easily extend to the former one as well.
}

\hide{
\begin{figure}[t]
    \centering
    \includegraphics[width=0.7\columnwidth]{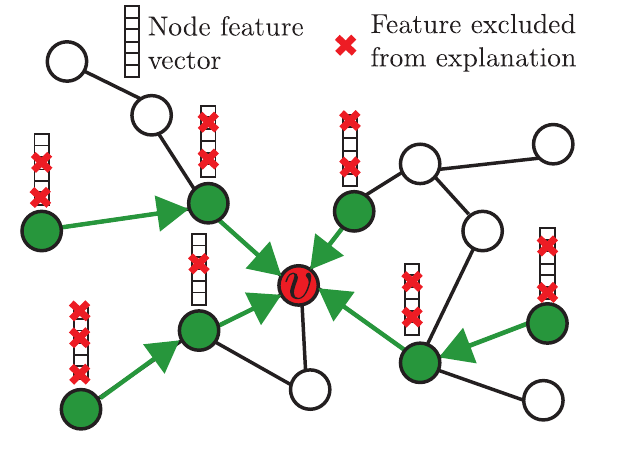}
    \vspace{-4mm}
    \caption{For $v$'s explanation $G_S$ (in green), \name identifies what feature dimensions of $G_S$'s nodes are essential for prediction at $v$
    by learning a node feature mask $M_T$.}
    \label{fig:including-node-features}
    \vspace{-5mm}
\end{figure}
}

To identify what node features are most important for prediction $\hat{y}$, \name learns a feature selector $F$ for nodes in explanation $G_S$. Instead of defining $X_S$ to consists of all node features, \ie, $X_S = \{x_j | v_j \in G_S\}$, \name considers $X_S^F$ as a subset of features of nodes in $G_S$, which are defined through a binary feature selector $F \in \{0,1\}^d$ (Figure~\ref{fig:definition-node-features}B):
%
\begin{equation}
    X_S^F = \{x_{j}^F | v_j \in G_S\}, \quad x_j^F = [x_{j, t_1}, \ldots, x_{j, t_k}] \text{~for~} F_{t_i} = 1,
\end{equation}
where $x_j^F$ has node features that are not masked out by $F$. 
Explanation $(G_S, X_S)$ is then jointly optimized for maximizing the mutual information objective:
\begin{equation}
    \label{eq:mi_obj_feat}
    \max_{G_S, F} MI\left(Y, (G_S, F)\right) = H(Y) - H(Y | G = G_S, X = X_S^F),
\end{equation}
which represents a modified objective function from Eq.~(\ref{eq:mi_obj}) that considers structural and node feature information to generate an explanation for prediction $\hat{y}$.

\xhdr{Learning binary feature selector $F$}
%
We specify $X_S^F$ as $X_S \odot F$, where $F$ acts as a feature mask that we need to learn. Intuitively, if a particular feature is not important, the corresponding weights in GNN's weight matrix take values close to zero. In effect, this implies that masking the feature out does not decrease predicted probability for $\hat{y}.$ Conversely, if the feature is important then masking it out would decrease predicted probability.
However, in some cases this approach ignores features that are important for prediction but take values close to zero.  
To address this issue we marginalize over all feature subsets and use a Monte Carlo estimate to sample from empirical marginal distribution for nodes in $X_S$ during training~\citep{zintgraf2017visualizing}. Further, we use a reparametrization trick~\citep{kingma2013auto} to backpropagate gradients in Eq.~(\ref{eq:mi_obj_feat}) to the feature mask $F$. In particular, to backpropagate through a $d$-dimensional random variable $X$ we reparametrize $X$ as: $X = Z + (X_S - Z) \odot F$ s.t. $\sum_j {F}_j \le K_F$, where $Z$ is a $d$-dimensional random variable sampled from the empirical distribution and $K_F$ is a parameter representing the maximum number of features to be kept in the explanation. 


%
\hide{
\begin{equation}
    \label{eq:marg_missing_feat}
    P_\Phi(Y | G_S, X_S^T) = \sum_{x_{i,t_k} \in D\backslash T} P(x_{i,t_k}) P_\Phi(Y | G=G_S, X=\{X_S^T, x_{i,t_k}\}).
\end{equation}
Here, $X = \{X_S^T, x_{i,t_k}\}$ is a random variable due to the missing features $x_j$ \marinka{Missing feature vs.  zero-valued features vs. constant feature?}, whose distribution is approximated with the empirical distribution from data.
Rather than computing an importance score for each feature dimension, we instead want to incorporate features into the objective in Eq.~(\ref{eq:mi_obj_feat}). 
The key challenge here is how to perform backpropagation through a random variable $X$, which we address with 
Importantly, this formulation allows us to backpropagate gradients from \name's objective to the feature mask $M_T$.
}

\hide{
Rather than computing an importance score for each feature dimension, we instead want to incorporate features into the objective in Eq.~(\ref{eq:mi_obj_feat}). 
The key challenge here is how to perform backpropagation through a random variable $X$, which we address with a reparametrization trick~\citep{kingma2013auto} for representing $X$ in Equation \ref{eq:marg_missing_feat}:
\begin{equation}
    \label{eq:missing_feat_reparam}
    X = Z + (X_S - Z) \odot M_T, \quad \textrm{s.t.} \quad \sum_j {M_T}_j \le K,
\end{equation}
where $X_S$ denotes node features $\{x_j | v_j \in G_S\}$, $Z$ is a $n$-by-$d$-dimensional random variable sampled from the empirical marginal distribution of nodes $x_j \in X_S$ during training, and $K$ is the desired explanation size.
Importantly, this formulation allows us to backpropagate gradients from \name's objective to the feature mask $M_T$.
}
\hide{
To see how the reparametrization allows optimization of feature mask, consider the scenario when $i$-th dimension is not important; that is, any sample of $Z$ from the empirical marginal distribution does not affect \gnn's prediction. In this case, the constraint will pull the corresponding mask value $M_i$ towards $0$, and as a result, the random variable $Z$ is fed into the \gnn.
On the other hand, if $i$-th dimension is very important, the random sample of $Z$ will degrade \gnn's performance, resulting in a larger value of objective function. Therefore, to minimize the loss, the mask value $M_i$ will go towards $1$, when $X$ becomes the deterministic $X_S$, which are the actual node features.
}

\hide{
To learn a binary feature selector $T$ we use continuous relaxation of intractable combinatorial optimization.
We approximate $X_S^T$ by $X_S \odot M_T$, where $M_T$ is a feature selection mask matrix and needs to be estimated/learned. 
However, consider the case where some feature $x_i$ takes value $0$ but that value of zero provides useful information for \gnn's prediction.
In this case, the feature $x_i$ should be retained and not be masked out, but our approximation proposed above would ignore it and filter the feature $x_i$ out. However, if $x_i \neq 0$ and also being important for prediction, it would not be filtered out.
To resolve this issue with important zero-valued features we marginalize over all possible choices of features in $\{0,1\}^D \backslash T$. 
In practice, we use Monte Carlo estimate of the marginals of these missing features~\cite{zintgraf2017visualizing}:
\begin{equation}
    \label{eq:marg_missing_feat}
    P_\Phi(Y | G_S, X_S^F) = \sum_{x_i} P(x_i) P_\Phi(Y | G=G_S, X=\{X_S^F, x_i\}),
\end{equation}
where $X = \{X_S^T, x_i\}$ is a random variable due to the missing features $x_i$ whose distribution is approximated with the empirical distribution from data.
However, in \name, rather than computing an importance score for each feature dimension, we instead want to incorporate features into the objective in Equation~\ref{eq:mi_obj_feat}. 
The key challenge here is to perform backpropagation through a random variable $X$, which we address through a reparametrization trick for $X$:
\begin{equation}
    \label{eq:missing_feat_reparam}
    X = Z + (X_S - Z) \odot \sigma(F), \quad \textrm{s.t.} \quad \sum_j \sigma({F}_j) \le k
\end{equation}
Here $X_S$ is still defined as the deterministic node features $\{x_i | v_i \in G_S\}$.
Such a formulation allows the backpropagation of gradients from the objective to the feature mask $M_T$.

During training, a random variable $Z$ of dimension $n$-by-$d$ is sampled from the empirical marginal distribution of all nodes $x_i \in X_S$.
To see how the reparametrization allows optimization of feature mask, consider the scenario when $i$-th dimension is not important; that is, any sample of $Z$ from the empirical marginal distribution does not affect \gnn's prediction. In this case, the constraint will pull the corresponding mask value $M_i$ towards $0$, and as a result, the random variable $Z$ is fed into the \gnn.
On the other hand, if $i$-th dimension is very important, the random sample of $Z$ will degrade \gnn's performance, resulting in a larger value of objective function. Therefore, to minimize the loss, the mask value $M_i$ will go towards $1$, when $X$ becomes the deterministic $X_S$, which are the actual node features.
}



\xhdr{Integrating additional constraints into explanations}
To impose further properties on the explanation we can extend \name's objective function in Eq.~(\ref{eq:mi_obj_feat}) with regularization terms. 
For example, we use element-wise entropy to encourage structural and node feature masks to be discrete. Further, \name can encode domain-specific constraints through techniques like Lagrange multiplier of constraints or additional regularization terms. 
We include a number of regularization terms to produce explanations with desired properties.
We penalize large size of the explanation by adding the sum of all elements of the mask paramters as the regularization term.

Finally, it is important to note that each explanation must be a valid computation graph. In particular, explanation $(G_S, X_S)$ needs to allow GNN's neural messages to flow towards node $v$ such that GNN can make prediction $\hat{y}$. 
Importantly, \name automatically provides explanations that represent valid computation graphs because it optimizes structural masks across entire computation graphs. Even if a disconnected edge is important for neural message-passing, it will not be selected for explanation as it cannot influence GNN's prediction. In effect, this implies that the explanation $G_S$ tends to be a small connected subgraph.

\hide{
\begin{enumerate}[leftmargin=*]
    \item \textbf{Explanation size:} Explanations that are too complex often do not give useful insights. We regularize the size of explanations $|G_S| = \sum_{i,j} {A_S}_{i,j} \le k$. 
    \item \textbf{Discrete structural and feature masks:} 
    We regularize the element-wise entropy of structural (feature) mask $M$ ($M_T$). 
    \item \textbf{Application-specific prior knowledge:} We include app\-lica\-tion-specific prior knowledge using Laplacian regularization.
    For example, in graphs that exhibit homophily, informative messages flow between nodes with similar labels. We can encourage such prior via a quadratic form regularization $f^T L_S f$, where $f$ is the label function on each node, and $L_S = D_S - A_S$ is the graph Laplacian corresponding to adjacency matrix $A_S$.
    \item \textbf{Valid \gnn computation graph:} The explanation of $v_i$'s prediction should be a subgraph of $v_i$'s computation graph that is itself a valid computation graph. In particular, the explanation needs to allow message passing towards the center node $v_i$ in order to generate node prediction $\hat{y}$. In practice, we find that this property is naturally achieved in \name without explicit regularization. This is because \name jointly optimizes the mask across all edges. It encourages finding computation graphs that can pass information to the center node $v_i$ and discourages disconnected edges (even if a disconnected edge is important for message-passing, it will not be selected if it cannot influence the center node).
\end{enumerate}
}

\hide{
\subsection{Joint consideration of structural and node feature information in explanations}
\label{sec:incorporate-features}


\hide{
Node feature information plays an important role in computing messages between nodes. The explanations therefore should take into account the feature information. However, in a \gnn node feature information is propagated over multiple layers, which means important features could differ between the nodes and the features might interact in non-linear ways across the multi-layer \gnn. We tackle this problem by learning a feature mask (one for the entire \gnn or one per node in the \gnn computation graph). The mask then acts as a feature selector and also prevents the feature information from ``leaking'' through the layers of the \gnn.
}


\hide{
We explicitly consider what feature dimensions are essential for message passing, \emph{i.e.} feature selection for nodes in $G_S$. 
In addition to optimizing Eq.~(\ref{eq:ent_obj}) over $G_S$, we also perform optimization that selects important features in $X_S$. 
Two approaches are possible: (1) we can optimize the essential dimensions for each node's feature $x_i \in G_S$ separately, or (2) we can optimize a single set of important dimensions for all nodes' features involved in the \gnn computation graph.
The former allows for finer granularity of feature importance but provides more complex explanations, while the latter is computationally more efficient and provides more concise explanations. Here we adopt the latter approach, but the following analysis can easily extend to the former one as well.
}

\hide{
\begin{figure}[t]
    \centering
    \includegraphics[width=0.7\columnwidth]{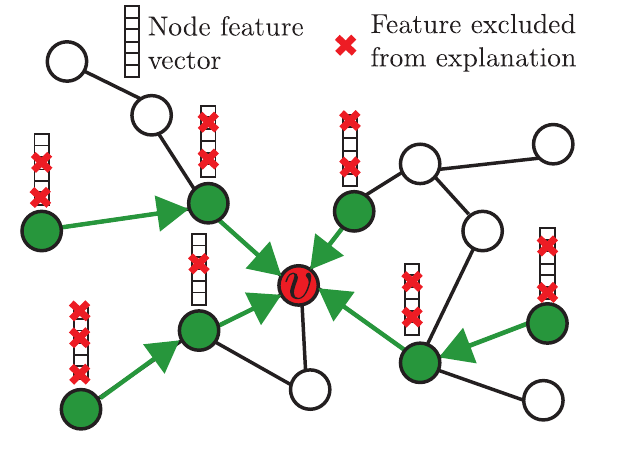}
    \vspace{-4mm}
    \caption{For $v$'s explanation $G_S$ (in green), \name identifies what feature dimensions of $G_S$'s nodes are essential for prediction at $v$
    by learning a node feature mask $M_T$.}
    \label{fig:including-node-features}
    \vspace{-5mm}
\end{figure}
}

To identify what dimensions of node feature information are important for prediction, \name also learns a binary selector for the features. In particular, for a subgraph $G_S$, instead of defining $X_S$ to consists of all node features, \ie, $X_S = \{x_j | v_j \in G_S\}$, \name defines $X_S^T$ as a subset of features of nodes in $G_S$ through a binary feature selector $T \in \{0,1\}^D$ (Fig.~\ref{fig:definition-node-features}, Right):
\begin{align}
    X_S^T = \{x_{j}^T | v_j \in S\}, \;\;\;\; x_j^T = [x_{j, t_1}, \ldots, x_{i, t_k}], \text{where~} t_k \in T.
\end{align}
Explanation $(G_S, X_S)$ is then jointly optimized for maximizing the mutual information objective:
\begin{equation}
    \label{eq:mi_obj_feat}
    \max_{G_S, T} MI\left(Y, (G_S, T)\right) = H(Y) - H(Y | G = G_S, X = X_S^T)
\end{equation}
that represents a modified objective function from Eq.~(\ref{eq:mi_obj}) that considers structural and node feature information when learning an explanation for GNN's prediction $\hat{y}$.

\xhdr{Learning binary feature selectors}
%
To learn a binary feature selector $T$ we approximate $X_S^T$ by $X_S \odot M_T$, where $M_T$ is a feature mask that we need to learn. 
The main challenge here is to consider the case when a particular feature takes value $0$ \marinka{What if the feature is constant?} but that feature is important for \gnn's prediction.
That feature clearly not be masked out. 
To address the challenge when zero-valued features are relevant for prediction, we marginalize over all possible choices of features in $\{0,1\}^{D \backslash T}$ using a Monte Carlo estimate of the marginals~\citep{zintgraf2017visualizing} \marinka{Is $D$ the number of node features? We use $d$ in some other parts. Also, is $Z$ an $n$-by-$d$-dimensional varible or is it $n$-by-$D$-dimensional (see below, just after eq 9)?}: \rex{D is the set $\{1,2,\ldots,d\}$ see slack}
\begin{equation}
    \label{eq:marg_missing_feat}
    P_\Phi(Y | G_S, X_S^T) = \sum_{x_j \in D\backslash T} P(x_j) P_\Phi(Y | G=G_S, X=\{X_S^T, x_j\}).
\end{equation}
Here, $X = \{X_S^T, x_j\}$ is a random variable due to the missing features $x_j$ \marinka{Missing feature vs.  zero-valued features vs. constant feature?}, whose distribution is approximated with the empirical distribution from data.

Rather than computing an importance score for each feature dimension, we instead want to incorporate features into the objective in Eq.~(\ref{eq:mi_obj_feat}). 
The key challenge here is how to perform backpropagation through a random variable $X$, which we address with a reparametrization trick~\citep{kingma2013auto} for representing $X$ in Equation \ref{eq:marg_missing_feat}:
\begin{equation}
    \label{eq:missing_feat_reparam}
    X = Z + (X_S - Z) \odot M_T, \quad \textrm{s.t.} \quad \sum_j {M_T}_j \le K,
\end{equation}
where $X_S$ denotes node features $\{x_j | v_j \in G_S\}$, $Z$ is a $n$-by-$d$-dimensional random variable sampled from the empirical marginal distribution of nodes $x_j \in X_S$ during training, and $K$ is the desired explanation size.
Importantly, this formulation allows us to backpropagate gradients from \name's objective to the feature mask $M_T$.
\hide{
To see how the reparametrization allows optimization of feature mask, consider the scenario when $i$-th dimension is not important; that is, any sample of $Z$ from the empirical marginal distribution does not affect \gnn's prediction. In this case, the constraint will pull the corresponding mask value $M_i$ towards $0$, and as a result, the random variable $Z$ is fed into the \gnn.
On the other hand, if $i$-th dimension is very important, the random sample of $Z$ will degrade \gnn's performance, resulting in a larger value of objective function. Therefore, to minimize the loss, the mask value $M_i$ will go towards $1$, when $X$ becomes the deterministic $X_S$, which are the actual node features.
}

\hide{
To learn a binary feature selector $T$ we use continuous relaxation of intractable combinatorial optimization.
We approximate $X_S^T$ by $X_S \odot M_T$, where $M_T$ is a feature selection mask matrix and needs to be estimated/learned. 
However, consider the case where some feature $x_i$ takes value $0$ but that value of zero provides useful information for \gnn's prediction.
In this case, the feature $x_i$ should be retained and not be masked out, but our approximation proposed above would ignore it and filter the feature $x_i$ out. However, if $x_i \neq 0$ and also being important for prediction, it would not be filtered out.
To resolve this issue with important zero-valued features we marginalize over all possible choices of features in $\{0,1\}^D \backslash T$. 
In practice, we use Monte Carlo estimate of the marginals of these missing features~\cite{zintgraf2017visualizing}:
\begin{equation}
    \label{eq:marg_missing_feat}
    P_\Phi(Y | G_S, X_S^T) = \sum_{x_i \in D\backslash T} P(x_i) P_\Phi(Y | G=G_S, X=\{X_S^T, x_i\}),
\end{equation}
where $X = \{X_S^T, x_i\}$ is a random variable due to the missing features $x_i$ whose distribution is approximated with the empirical distribution from data.
However, in \name, rather than computing an importance score for each feature dimension, we instead want to incorporate features into the objective in Equation~\ref{eq:mi_obj_feat}. 
The key challenge here is to perform backpropagation through a random variable $X$, which we address through a reparametrization trick for $X$:
\begin{equation}
    \label{eq:missing_feat_reparam}
    X = Z + (X_S - Z) \odot M_T, \quad \textrm{s.t.} \quad \sum_j {M_T}_j \le k
\end{equation}
Here $X_S$ is still defined as the deterministic node features $\{x_i | v_i \in G_S\}$.
Such a formulation allows the backpropagation of gradients from the objective to the feature mask $M_T$.

During training, a random variable $Z$ of dimension $n$-by-$d$ is sampled from the empirical marginal distribution of all nodes $x_i \in X_S$.
To see how the reparametrization allows optimization of feature mask, consider the scenario when $i$-th dimension is not important; that is, any sample of $Z$ from the empirical marginal distribution does not affect \gnn's prediction. In this case, the constraint will pull the corresponding mask value $M_i$ towards $0$, and as a result, the random variable $Z$ is fed into the \gnn.
On the other hand, if $i$-th dimension is very important, the random sample of $Z$ will degrade \gnn's performance, resulting in a larger value of objective function. Therefore, to minimize the loss, the mask value $M_i$ will go towards $1$, when $X$ becomes the deterministic $X_S$, which are the actual node features.
}



\xhdr{Integrating additional constraints into explanations}
To impose additional desired properties on the explanation we can extend \name's objective function in Eq.~(\ref{eq:opt_node_class}) with regularization terms. We regularize explanation size as: $|G_S| = \sum_{j,k} {A_S}_{j,k} \le K$, and encourage structural (feature) mask $M$ ($M_T$) to be discrete via the element-wise entropy. Further, \name can encode domain-specific prior information through techniques like Laplacian regularization. 
Finally, it is important to note that each explanation must be a valid computation graph itself. In particular, explanation needs to allow GNN's neural messages to flow towards node $v_i$ in order to generate prediction $\hat{y}$. 
\name achieves this property without explicit regularization because it optimizes a structural mask across entire computation graph, which discourages disconnected edges in the explanation. This is because even if a disconnected edge is important for neural message-passing, it will not be selected as it cannot influence $\hat{y}$.

\hide{
\begin{enumerate}[leftmargin=*]
    \item \textbf{Explanation size:} Explanations that are too complex often do not give useful insights. We regularize the size of explanations $|G_S| = \sum_{i,j} {A_S}_{i,j} \le k$. 
    \item \textbf{Discrete structural and feature masks:} 
    We regularize the element-wise entropy of structural (feature) mask $M$ ($M_T$). 
    \item \textbf{Application-specific prior knowledge:} We include app\-lica\-tion-specific prior knowledge using Laplacian regularization.
    For example, in graphs that exhibit homophily, informative messages flow between nodes with similar labels. We can encourage such prior via a quadratic form regularization $f^T L_S f$, where $f$ is the label function on each node, and $L_S = D_S - A_S$ is the graph Laplacian corresponding to adjacency matrix $A_S$.
    \item \textbf{Valid \gnn computation graph:} The explanation of $v_i$'s prediction should be a subgraph of $v_i$'s computation graph that is itself a valid computation graph. In particular, the explanation needs to allow message passing towards the center node $v_i$ in order to generate node prediction $\hat{y}$. In practice, we find that this property is naturally achieved in \name without explicit regularization. This is because \name jointly optimizes the mask across all edges. It encourages finding computation graphs that can pass information to the center node $v_i$ and discourages disconnected edges (even if a disconnected edge is important for message-passing, it will not be selected if it cannot influence the center node).
\end{enumerate}
}

}

\subsection{Multi-instance explanations through graph prototypes}
\label{sec:global-model}

\hide{
\jure{I would write a short paragraph(s) (say 1/2 page) saying something like:
(1) Problem intro and definition
(2) Why it is hard/challenging (mapping the explainer graphs)
3) Overview of our solution
4) Point to appendix for more information.
}
}

The output of a single-instance explanation (Sections \ref{sec:local-model} and \ref{sec:incorporate-features}) is a small subgraph of the input graph and a small subset of associated node features that are most influential for a single prediction. To answer questions like ``How did a GNN predict that a given set of nodes all have label $c$?'', we need to obtain a global explanation of class $c$. Our goal here is to provide insight into how the identified subgraph for a particular node relates to a graph structure that explains an entire class. \name can provide multi-instance explanations based on graph alignments and prototypes. Our approach has two stages: 

First, for a given class $c$ (or, any set of predictions that we want to explain), we first choose a reference node $v_c$, for example, by computing the mean embedding of all nodes assigned to $c$.
%
%
We then take explanation $G_S(v_c)$ for reference $v_c$ and align it to explanations of other nodes assigned to class $c$. 
Finding optimal matching of large graphs is challenging in practice. 
However, the single-instance \name generates small graphs (Section~\ref{sec:incorporate-features}) and thus near-optimal pairwise graph matchings can be efficiently computed.

Second, we aggregate aligned adjacency matrices into a graph prototype $A_\textrm{proto}$ using, for example, a robust median-based approach.
Prototype $A_{\mathrm{proto}}$ gives insights into graph patterns shared between nodes that belong to the same class. One can then study prediction for a particular node by comparing explanation for that node's prediction (\ie,  returned by single-instance explanation approach) to the prototype (see Appendix for more information).

%
\hide{
Let $A_v$ and $X_v$ be the adjacency matrix and the associated feature matrix of the to-be-aligned computation subgraph. Similarly let $A^*$ be the adjacency matrix and associated feature matrix of the reference computation subgraph.
Then we optimize the relaxed alignment matrix $P \in \mathbb{R}^{n_v \times n^*}$, where $n_v$ is the number of nodes in $G_S(v)$, and $n^*$ is the number of nodes in $G_S(v_c)$ as follows:
\begin{equation}
    \label{eq:glob}
    \min_P | P^T A_v P - A^*| + |P^T X_v - X^*|.
\end{equation}
The first term in Equation~\ref{eq:glob} specifies that after alignment, the aligned adjacency for $G_S(v)$ should be as close to $A^*$ as possible. The second term Equation~\ref{eq:glob} specifies that the features should for the aligned nodes should also be close. 

%
%
%
%
This procedure gives us adjacency matrices $A_S$ for all nodes predicted to belong to $c$, where the matrices are aligned with respect to the node ordering in the reference $v_c$'s matrix $A_S(v_c)$. 
}

\hide{
The output of a single-instance \name indicates what graph structural and node feature information  is important for a given prediction. To obtain an understanding of ``why is a given set of nodes classified with label $y$'', we want to also obtain a global explanation of the class, which can shed light on how the identified structure for a given node is related to a prototypical structure unique for its label. To this end, we propose an alignment-based multi-instance \name. 

For any given class, we first choose a reference node. Intuitively, this node should be a prototypical node for the class. Such node can be found by computing the mean of the embeddings of all nodes in the class, and choose the node whose embedding is the closest to the mean. Alternatively, if one has prior knowledge about the important computation subgraph, one can choose one which matches most to the prior knowledge.

Given the reference node for class $c$, $v_c$, and its associated important computation subgraph $G_S(v_c)$, 
we align each of the identified computation subgraphs for all nodes in class $c$ to the reference $G_S(v_c)$.
Utilizing the idea in the context of differentiable pooling~\cite{ying2018hierarchical}, we use the a relaxed alignment matrix to find correspondence between nodes in an computation subgraph $G_S(v)$ and nodes in the reference computation subgraph $G_S(v_c)$.
Let $A_v$ and $X_v$ be the adjacency matrix and the associated feature matrix of the to-be-aligned computation subgraph. Similarly let $A^*$ be the adjacency matrix and associated feature matrix of the reference computation subgraph.
Then we optimize the relaxed alignment matrix $P \in \mathbb{R}^{n_v \times n^*}$, where $n_v$ is the number of nodes in $G_S(v)$, and $n^*$ is the number of nodes in $G_S(v_c)$ as follows:
\begin{equation}
    \label{eq:glob}
    \min_P | P^T A_v P - A^*| + |P^T X_v - X^*|.
\end{equation}
The first term in Equation~\ref{eq:glob} specifies that after alignment, the aligned adjacency for $G_S(v)$ should be as close to $A^*$ as possible. The second term Equation~\ref{eq:glob} specifies that the features should for the aligned nodes should also be close. 

In practice, it is often non-trivial for the relaxed graph matching to find a good optimum for matching 2 large graphs. 
However, thanks to the single-instance explainer, which produces concise subgraphs for important message-passing, a matching that is close to the best alignment can be efficiently computed.



\xhdr{Prototype by alignment}
We align the adjacency matrices of all nodes in class $c$, such that they are aligned with respect to the ordering defined by the reference adjacency matrix. We then use median to generate a prototype that is resistent to outliers,
$A_{\mathrm{proto}} = \mathrm{median}( A_i)$, where $A_i$ is the aligned adjacency matrix representing explanation for $i$-th node in class $c$.
Prototype $A_{\mathrm{proto}}$ allows users to gain insights into structural graph patterns shared between nodes that belong to the same class. Users can then investigate a particular node by comparing its explanation to the class prototype.
}

\subsection{\name model extensions}
\label{sec:other-tasks}


\xhdr{Any machine learning task on graphs}
In addition to explaining node classification, \name provides explanations for link prediction and graph classification with no change to its optimization algorithm. When predicting a link $(v_j, v_k)$, \name learns two masks $X_S(v_j)$ and $X_S(v_k)$ for both endpoints of the link. When classifying a graph, the adjacency matrix in Eq.~(\ref{eq:opt_node_class}) is the union of adjacency matrices for all nodes in the graph whose label we want to explain.
However, note that in graph classification, unlike node classification, due to the aggregation of node embeddings, it is no longer true that the explanation $G_S$ is necessarily a connected subgraph. Depending on application, in some scenarios such as chemistry where explanation is a functional group and should be connected, one can extract the largest connected component as the explanation.

\hide{
In addition to node classification, \name can provide explanations for link prediction and graph classification tasks. For example,  we modify the \name objective function in Equation \ref{eq:opt_general} as follows.

When predicting a link between nodes $v_i$ and $v_j$, $Y$ represents a distribution of existence of links. Quantity $P_\Phi$ in Equation \ref{eq:opt_general} is thus:
\begin{equation}
    P_\Phi(Y | G_1 = A_{S_i} \odot M_1, G_2 = A_{S_j} \odot M_2, X=X_S),
\end{equation}
meaning that \name learns two masks that identify important computation subgraphs for both nodes involved in the predicted link. When classifying graphs, $Y$ represents a distribution of label classes for graphs. This means that quantity $A_S$ in \name objective function is no longer an adjacency matrix for the computation graph of a particular node $v_i$. Instead, it is the union of computation graphs across all nodes in the graph whose graph-level label we want to explain. Importantly, optimization procedure for explaining link prediction and graph classification is the same as that for node classification.
}

\xhdr{Any GNN model}
Modern GNNs are based on message passing architectures on the input graph.
The message passing computation graphs can be composed in many different ways and \name can account for all of them. Thus, \name can be applied to: Graph Convolutional Networks~\citep{kipf2016semi}, Gated Graph Sequence Neural Networks~\citep{li2015gated}, Jumping Knowledge Networks~\citep{xujumping}, Attention Networks~\citep{velickovic2018graph}, Graph Networks~\citep{battaglia}, GNNs with various node aggregation schemes~\citep{fastgcn, Chen2018StochasticTO,Huang2018AdaptiveST, graphsage,ying2018hierarchical,pinsage,xu2018powerful}, Line-Graph NNs~\citep{chen2018supervised}, position-aware GNN~\citep{You2019position}, and many other GNN architectures.


\hide{
\jure{I would make this paragraph more a laundry list of methods without much explanation.
The message passing computation graph can be composed in many different ways and \name can account for them all. Thus, \name can be applied to: NAME (CITE), NAME (CITE), ...., and many other GNN architectures.}

\name is GNN-model-agnostic and only accesses a trained GNN model in terms of forward and backward propagation. Since it generates explanations by learning masks for computation graphs, it can explain predictions made by any GNN. In general, the computation graph $G_c(v_i)$ for node $v_i$ is obtained by performing a graph traversal of depth $L$, \eg, a Breadth-First Search (BFS), using $\mathcal{N}_{v_i}$ as the neighborhood definition. For example, in Graph Convolutional Networks~\citep{kipf2016semi}, the computation graph $G_c(v_i)$ is simply an $L$-hop neighborhood of $v_i$ in the input graph $G$. However, in other \gnns, such as Jumping Knowledge Networks~\citep{xujumping}, Attention Networks~\citep{velickovic2018graph}, Graph Networks~\citep{battaglia}, GNNs with node aggregation~\citep{hamilton2017representation}, and Line-Graph NNs~\citep{chen2018supervised}, computation graphs are different from exhaustive $L$-hop neighborhoods. \rex{As an example, for the experiments, we also run our method to explain GAT, and show in appendix?}
}




\xhdr{Computational complexity}
The number of parameters in \name's optimization depends on the size of computation graph $G_c$ for node $v$ whose prediction we aim to explain. In particular, $G_c(v)$'s adjacency matrix $A_c(v)$ is equal to the size of the mask $M$, which needs to be learned by \name. However, since computation graphs are typically relatively small, compared to the size of exhaustive $L$-hop neighborhoods (\eg, 2-3 hop neighborhoods~\citep{kipf2016semi}, sampling-based neighborhoods~\citep{pinsage}, neighborhoods with attention~\citep{velickovic2018graph}), \name can effectively generate explanations even when input graphs are large.

\hide{\subsection{Overview of \longname}

\marinka{
The information that a \gnn~ relies on for computing $\mathbf{z}_i$ can be completely described by its computation graph $G_c(v_i)$ and the associated feature set $X_c(v_i) = \{x_j | v_j \in G_c(v_i)\}.$
%
We thus use these elements as inputs to the \gnn~model $\hat{y} = \Phi(G_c(v_i), X_c(v_i))$.
Therefore, when constructing a local explanation for the prediction a \gnn has made on $v_i$, we only consider structural information present in $G_c(v_i)$, and node features from $X_c(v_i)$.
}

We introduce GNN-Explainer, a tool for post-hoc interpretation of predictions generated by a pre-trained \gnn. The approach is able to identify nodes that are locally relevant to a given prediction, similarly to existing interpretation methods. However, it is also able to identify relevant structures in the induced computation graph, which is a sample of a given node's neighborhood. We also present an extension to provide insight into global structures in the graph that are relevant to a given class of nodes. 

To achieve these goals, we define a useful \gnn~ explanation as being composed of two key components: a local explainer (\lexp) and a global explainer (\gexp). The local explainer \lexp~ identifies, among the computation graph $G_c(v_i)$ of each node $v_i$, the particular feature dimensions and message-passing pathways that most contribute to the prediction made by the model for the given node's label.

In other words, we aim to identify a subgraph of the computation graph $G_c(v_i)$, among all possible subgraphs of $G_c(v_i)$, and a subset of feature dimensions for node features, that is the most influential in the model's prediction.
In this process both important message-passing and node features are jointly determined, due to the message aggregation scheme of \gnn s, which incorporates both aspects in a single forward pass.

In a second step, the global explainer \gexp~ will aggregate, in a class-specific way, information from the local explainer \lexp. This way, the local explanations are summarized into a prototype computation graph. The global explainer \gexp~ therefore allows the comparison of different computation graphs across different nodes in graphs, offering insights into the distribution of computation graphs for nodes in a certain label class.

\marinka{ but \name can be easily adapted to the task of link prediction and graph classification.}

\rex{How to use explainer}

\subsection{Local model}
In a classification setting, the output of the \gnn~ is a distribution $P_\Phi(Y | G_c, X_c)$, where $Y$ is the random variable that takes the value of the labels $\{1, \ldots, C\}$, indicating the probability of the node belonging to each of the $C$ classes, as predicted by the \gnn~ model.

In the local model \lexp, we identify a subgraph $G_S \subseteq G_c(v_i)$ with the associated features $X_S = \{\mathbf{x}_i | i \in G_S\}$ which is important for the \gnn~ prediction of $v_i$. 
We define this notion of importance via mutual information in equation~\ref{eq:mi_obj}:
\begin{equation}
    \label{eq:mi_obj}
    \max_{G_S } MI\left(Y, (G_S, X_S)\right) = H(Y) - H(Y | G=G_S, X=X_S).
\end{equation}
In addition to the constraints $G_S \subseteq G_c(v_i)$, $X_S = \{\mathbf{x}_i | i \in G_S\}$,
we also impose the size constraint $|G_S| \le k$, which specifies that the subgraph for explanation should be concise and its size should not exceed $k$ nodes.\dylan{add a note on how to be concise with few nodes or few edges?}

In our post-hoc scenario, the model $\Phi$ is fixed, hence the entropy of the label confidence $H(Y)$ is also fixed. Therefore, maximizing of the mutual information between the predicted label distribution and the computation subgraph with its features $(G_S, X_S)$ is equivalent to the minimization of the conditional entropy $H(Y | G=G_S, X=X_S)$, which can be expressed as follows:
\begin{equation}
    \label{eq:ent_obj}
    H(Y | G=G_S, X=X_S) = -\mathbb{E}_{Y | (G_S, X_S)} \left[ \log P_\Phi (Y | G=G_S, X=X_S) \right]
\end{equation}

In other words, the most important subgraph $G_S$ and its associated features $X_S$ will minimize the uncertainty predicted by the \gnn~ model $\Phi$ by performing its message-passing according to the computation graph defined by $G_S$. In the task of classification, this corresponds to maximizing the class probability of the predicted class $\hat{y}$, where $\hat{y} = \argmax_{y \in Y} P_\Phi(Y | G_c, X_c)$.

The objective of maximizing mutual information has been explored in Deep Infomax~\cite{hjelm2018learning} and Deep Graph Infomax~\cite{Velickovic2018DeepGI}, in the context of learning local representations that correlate most with graph representations.

\xhdr{Motivation}
Intuitively, the local model \lexp sees edges in the computation graphs as having different roles.
Some edges, in combination, form important message-passing pathways, which allows useful features and hidden representations to be aggregated to the target node $v_i$ for prediction.

Other edges in the computation graph, however, could play an inhibitory role. The messages passed from these edges do not provide any additional information, and might even hurt the model performance. Indeed, the aggregator in Equation~\ref{eq:agg} needs to aggregate such uninformative messages together with the useful ones, which can thus dilute the signal.

The objective defined by \lexp aims to remove the inhibitory edges, in a sense denoising the computation graph to keep at most $k$ edges without compromising the mutual information between the computation graph and the model prediction. 

In practice, many datasets and tasks support such motivation. 
As illustrated in Figure TODO.
\rex{Figure illustration: important substructure embedded in a comp graph with many other connections}

\xhdr{Connection to Feature Selection}
The motivation for \lexp is also similar to feature selection explanation~\cite{Erhan2009VisualizingHF, chen2018learning, sundararajan_axiomatic_nodate, lundberg_unified_2017}, whose aim is to select a subset of feature dimensions that are most informative to the model prediction. However, an important distinction is that the importance of feature dimensions can usually be treated as independent: the importance of one dimension does not depend on the importance of the other.\dylan{The dependency is across nodes, not really feature dimensions, right?}
In contrast, the importance of edges in \gnn~ subgraphs are correlated.
A simple example is that: an edge could be important when there are another alternative pathway that forms a cycle, indicating a special structural role of the node to be predicted. Hence the decision to include this edge in $G_S$ depends on whether the alternate pathway is included. \rex{do we need figure to illustrate?}

Furthermore, in the feature selection setting, the feature dimension of different data points is fixed.
However, since the computation graph $G_c(v_i)$ is different for explanation of different node $v_i$, the choice of $G_S$ for each node $v_i$ does not have an exact correspondence. 
As a consequence, the optimization of the \lexp objective has to be local and specific to the individual computation graph $G_c(v_i)$.

\xhdr{Model optimization}
Direct optimization of the objective \ref{eq:mi_obj} is not tractable, since there are exponentially many discrete structures $G_S \subseteq G_c(v_i)$. 
We thus use an approximation similar to soft-attention~\CITE. Rather than optimizing $G_S$'s adjacency matrix $A_S \in \{0,1\}^{n \times n}$ as a discrete structure where $n$ is the size of the computation graph $G_c$, we allow fractional adjacency matrix for subgraph $G_S$: $A_S \in [0,1]^{n\times n}$ \footnote{For typed edges, we define $G_S \in [0,1]^{C_e \times n \times n}$, where $C_e$ is the number of classes.}.
To enforce the subgraph constraint,
the relaxed adjacency $A_S$ is subject to the constraint that $A_S[i,j] \le A_c[i, j], \forall i, j=1,\ldots,n$.
Such relaxation allows gradient descent to be performed on $G_S$.

The relaxed $G_S$ can be interpreted as a variational approximation of subgraph distributions of $G_c(v_i)$ using the mean field assumption. 
If we treat $G_S \sim \mathcal{G}$ as a random graph variable, then the objective \ref{eq:ent_obj} becomes:
\begin{equation}
    \label{eq:variational_relaxation}
    \min_{\mathcal{G}} \mathbb{E}_{G_S \sim \mathcal{G}} H(Y | G=G_S, X=X_S)
\end{equation}
When $\Phi$ is convex, by Jensen's inequality, Equation \ref{eq:variational_relaxation} has the upper bound which we optimize:
\begin{equation}
    \label{eq:variational_upper_bound}
    \min_{\mathcal{G}} H(Y | G=\mathbb{E}_{\mathcal{G}}[G_S], X=X_S)
\end{equation}
Using the mean field variational approximation, we decompose $\mathcal{G}$ into 
the multi-variate Bernoulli distribution $P_\mathcal{G}(G_S) = \prod_{(i,j) \in G_c(v_i)} {A_S}_{ij}$. Instead of optimizing $\mathbb{E}_{\mathcal{G}}[G_S]$, we optimize the expectation with respect to the mean-field approximation, namely $A_S$, whose $(i,j)$ entry is the expectation of the Bernoulli distribution of whether edge $(i,j)$ exists.

However, the graph neural network $\Phi$ with more than $1$ layer is generally not convex. Despite the non-convexity, in experiments we found that it often converges to a good local minima with help of regularizations that encourages discreteness.
This continuous approximation of adjacency matrix has also achieved success in the context of graph-based learning systems for adversarial attacks\CITE, neural architecture search\CITE, and differentiable pooling~\cite{ying2018hierarchical}.

To optimize the relaxed $A_S \in [0,1]^{n\times n}$, subject to the constraint $A_S[i,j] \le A_c[i, j], \forall i, j=1,\ldots,n$, we design a mask $M \in \mathbb{R}^{n\times n}$ as parameters to optimize. $A_S$ can be computed by taking element-wise multiplication of mask $\sigma(M)$ with $A_c$, and always obey the continuous relaxation of the subgraph constraint. Here $\sigma$ denotes the sigmoid function that maps $M$ to $[0,1]^{n\times n}$, but other activation functions that squash the input can be used.

Concretely, we summarize the objective of \lexp with the following optimization problem:
\begin{align}
    & \min_{M} -\mathbb{E}_{Y | A_S \odot \sigma(M), X_S} \left[ \log P_\Phi (Y | G=A_S \odot M, X=X_S) \right] \label{eq:opt_general} \\
    = & \min_{M}  -\sum_{c =1}^C \mathbbm{1}[y = c] \log P_\Phi (Y = y | G=A_S \odot \sigma(M), X=X_S) \label{eq:opt_node_class}
\end{align}
Here $\odot$ denotes element-wise multiplication.
In the forward propagation, $P_\Phi (Y | G=A_S \odot M, X=X_S)$ is computed by replacing the adjacency matrix $A_c$ for the computation graph with $A_S \odot M$ in the aggregation step \ref{eq:agg}. In another word, the mask controls how much information is passed for every edge.

An alternative to the continuous relaxation is to use Gumbel Softmax to allow learning of the discrete adjacency $A_S$ by its continuous approximation, which allows for backpropagation from the objective function. We empirically compare both approaches in the experiment section.

\xhdr{Obtaining subgraph from mask}
Once the mask is trained, we can either use a threshold to remove low values in the mask,
or rank the values of the mask as importance of each edge, and keep the top $K$ edges in the computation subgraph.


\subsection{Incorporation of features}
As explained in our desiderata, feature information plays an important role in computing messages between nodes. The explanation therefore should take into account the feature information.
In \lexp, features are considered in two ways.

Firstly, it is naturally incorporated in the objective function \ref{eq:opt_node_class}, since every subgraph $G_S$ has its different associated set of features $X_S$ which influences the model prediction $P_\Phi$. The importance of an edge not only depends on the structure of the computation graph, but also the features from which messages are generated.

Moreover, we specifically consider what feature dimensions are important for message passing, \emph{i.e.} feature selection for nodes in $G_S$. 
In addition to optimizing Equation \ref{eq:ent_obj} over all possible $G_S$, we also perform optimization on feature dimensions of $X_S$. 
Two approaches are possible: we can either optimize the important dimensions for each node' feature $x_i \in G_S$ separately, or we can optimize a single set of important dimensions for all nodes' features involved in the computation graph.
The former allows for finer granularity of feature importance, while the later is computationally more efficient to compute.
For simplicity of notation, we adopt the later approach, but the following analysis can extend to the former one as well.

For any given $G_S$, an identified important computational subgraph, 
instead of defining $X_S = \{x_i | v_i \in G_S$,
we define $\mathbb{X}$ as a family of feature vectors, with cardinality equal to that of the power set of feature dimensions $D = \{1, 2, \ldots, d\}$:
\begin{align}
    & \mathbb{X} = \{X_S^T | T \in 2^D\}, \text{where} \\
    & X_S^T = \{x_{i}^T | v_i \in S\}, x_i^T = [x_{i, t_1}, \ldots, x_{i, t_j}], \forall t_j \in T
\end{align}
Here the family $\mathbb{X}$ denotes all possible feature vectors obtained by removing a subset of dimensions in the original feature vectors in $G_S$.
\rex{refer to the figure, where some dimension of the feature vector is masked out}

Therefore in addition to optimizing the computational subgraph $G_S$, we also jointly optimize $T$, the subset of feature dimensions for each node in $G_S$ in an end-to-end way.
Maximization of the mutual information in Equation \ref{eq:mi_obj} is now:
\begin{equation}
    \label{eq:mi_obj_feat}
    \max_{G_S, T} MI\left(Y, (G_S, T)\right) = H(Y) - H(Y | G =G_S, X=X_S^T).
\end{equation}
Again, due to intractability of combinatorial optimization of $T$, we perform continuous relaxation.
In cases where $0$ indicates no information, $X_S^T$ can be approximated by $X_S \odot M_T$, where $M_T$ is a mask matrix on features to be optimized.

However, in cases where $0$ indicates a useful feature for a dimension, this approximation will lead to $0$ importance for that dimension. In such cases, we leverage the idea in feature selection to marginalize over all possible choices of the features in $2^D \backslash T$. In practice, we use Monte Carlo estimate of the marginals of these missing features\cite{zintgraf2017visualizing}.  
\begin{equation}
    \label{eq:marg_missing_feat}
    P_\Phi(Y | G_S, X_S^T) = \sum_{x_i \in D\backslash T} P(x_i) P_\Phi(Y | G=G_S, X=\{X_S^T, x_i\})
\end{equation}
$X = \{X_S^T, x_i\}$ now becomes a random variable due to the missing features $x_i$.
In practice, the true distribution of missing feature $x_i$ is unknown, but we can approximate with its empirical distribution from data.

However, in our setting, rather than computing an importance score for each feature dimension, we instead want to incorporate into the objective \ref{eq:mi_obj_feat}. 
To allow backpropagation through the random variable $X$, we utilize the reparametrization trick for parametrizing $X$:
\begin{equation}
    \label{eq:missing_feat_reparam}
    X = Z + (X_S - Z) \odot M_T, \quad \emph{s.t.} \sum_j {M_T}_j \le k
\end{equation}
Here $X_S$ is still the deterministic node features $\{x_i | v_i \in G_S\}$.
Such formulation allows backpropagation of gradients from objective to the feature mask $M_T$.

During training, random variable $Z$ of dimension $n$-by-$d$ is sampled from the empirical marginal distribution of all nodes $x_i \in X_S$.
To see why the reparametrization allows optimization of feature mask,
consiser the scenario when the $i$-th dimension is not important, which means that any sample of $Z$ from the empirical marginal distribution does not affect the prediction much. In this case, the constraint will pull the corresponding mask value $M_i$ towards $0$, and as a result, the random variable $Z$ is fed into the \gnn.
On the other hand, if the $i$-th dimension is very important, random sample of $Z$ will degrade the performance of $\gnn$, resulting in larger loss. Therefore, to minimize the loss, the mask value $M_i$ will go towards $1$, when $X$ becomes the deterministic $X_S$, which are the actual node features.

Furthermore, both approaches does not allow joint identification of important feature dimensions for nodes, and cannot naturally incorporate prior information of computation graph through regularization. \lexp takes into account of these two important components, which we address in the following section.


\xhdr{Incorporation of Prior Knowledge}
The advantage of the \lexp optimization framework is that the objective allows incorporation of our prior knowledge of desired form of the important computation subgraph.
For example, the following aspects are considered as side-objectives to optimize jointly with the objective in Equation \ref{eq:opt_node_class}.
\begin{enumerate}
    \item Size constraint: explanations that are too complex often do not provide insights. We therefore regularize the size of explanations $|G_S| = \sum_{i,j} {A_S}_{i,j} \le k$. In the case of continuous relaxation, we instead compute its size as $\sum_{i,j}\sigma(M_{i,j})$.
    In practice, instead of adding a constraint, we use a corresponding regularization term, with coefficient $
    \lambda_{\mathrm{size}}$. The user can control $\lambda_{\mathrm{size}}$ to incorporate the prior knowledge about the size of the explanation.
    \item We further encourage discreteness of the masks that we optimize, by regularizing the element-wise entropy of the masks $M$ and $M_T$. For example, the entropy of $M$ is computed by:
    \begin{equation}
        H(M) = -\sum_{1\le i,j \le n} \left( M_{i,j} \log M_{i,j} + (1-M_{i,j}) \log (1-M_{i,j}) \right)
    \end{equation}
    \item In some applications, we can further include priors such as Laplacian regularization\CITE.
    For example, in graphs that exhibit the property of homophily\CITE, information usually flows from nodes with similar labels. We can thus encourage such prior by the quadratic form regularization $f^T L_S f$, where $f$ is the label function on each node, and $L_S = D_S - A_S$ is the graph Laplacian corresponding to the adjacency $A_S$.
\end{enumerate}
The regularization terms are added to the final objective function to optimize the masks $M$ and $M_T$. 

Another desired property of explanation for node $v_i$ is that the resulting important computation subgraph $G_S$ needs to be a valid computation graph for computing node representations $\mathbf{h}_i$. 
In particular, the identified edge needs to allow message passing towards the center node $v_i$ to be predicted.
However, in practice we find that this prior knowledge is naturally incorporated without explicit regularization. 
Compared to gradient-based baselines which select edges with the largest gradient and could result in disconnectedness, in our framework, the mask for each edge is jointly optimized to achieve maximum certainty of model prediction. Therefore it would encourage computation graphs that can pass information to the center node, and discourage disconnected edges, because even these edges are important for message-passing, they will not be selected if the center node cannot reach these edges.

\xhdr{Output distillation}
\rex{Jiaxuan postprocessing}
\lexp provides a local explanation for a specific node $v_i$.
However, such explanation may not be fully convincing, as the \lexp may overfit to the local structure of the node.
Ideally, \lexp should output a local explanation that takes into account the prevalence of the explanation among all the nodes.
Therefore, we propose to distill the local explanation for each node, via extracting the common subgraph $G_S$ that is shared with other nodes within the same class label. 

Specifically, the distillation is achieved via \gnn s that have the representation power to approximately identify the isomorphism of $k$-hop neighbourhood graphs, where $k$ is the number of GNN layers \cite{DBLP:journals/corr/HamiltonYL17}.
We first extract the set of subgraphs $G_S \in \mathcal{G}_S$ for all the nodes with the same class label as $v_i$.
Then, we compute node embeddings for all the nodes in $\mathcal{G}_S$ via the trained GNN $\Phi$, resulting a node embedding matrix $Z$.
We then apply the DBSCAN clustering algorithm \cite{ester1996density}

\subsection{Global model}
\label{sec:global}
The output of local model \lexp indicates for each node, which structure is important for prediction of its label.
To obtain a deeper understanding of ``why it is classified as having label $y$'', we in addition want to obtain a global view of the class, which can shed light on how the identified structure for a given node is related to a prototypical structure unique for its label. To this end, we propose an alignment-based global model \gexp. 

For any given class, we first choose a reference node. Intuitively, this node should be a prototypical node for the class. Such node can be found by computing the mean of the embeddings of all nodes in the class, and choose the node whose embedding is the closest to the mean. Alternatively, if one has prior knowledge about the important computation subgraph, one can choose one which matches most to the prior knowledge.

Given the reference node for class $c$, $v_c$, and its associated important computation subgraph $G_S(v_c)$, 
\gexp aligns each of the identified computation subgraphs for all nodes in class $c$ to the reference $G_S(v_c)$.
Utilizing the idea in the context of differentiable pooling~\cite{ying2018hierarchical}, we use the a relaxed alignment matrix to find correspondence between nodes in an computation subgraph $G_S(v)$ and nodes in the reference computation subgraph $G_S(v_c)$.
Let $A_v$ and $X_v$ be the adjacency matrix and the associated feature matrix of the to-be-aligned computation subgraph. Similarly let $A^*$ be the adjacency matrix and associated feature matrix of the reference computation subgraph.
Then \gexp optimizes the relaxed alignment matrix $P \in \mathbb{R}^{n_v \times n^*}$, where $n_v$ is the number of nodes in $G_S(v)$, and $n^*$ is the number of nodes in $G_S(v_c)$, by Equation \ref{eq:glob}
\begin{equation}
    \label{eq:glob}
    \min_P | P^T A_v P - A^*| + |P^T X_v - X^*|
\end{equation}
\rex{maybe a small alignment figure. showing node correspondences. and there are some nodes that are not mapped to the reference. Maybe even better to show fractional alignment like diffpool}
The first term specifies that after alignment, the aligned adjacency for $G_S(v)$ should be as close to $A^*$ as possible. The second term specifies that the features should for the aligned nodes should also be close.

In practice, it is often non-trivial for the relaxed graph matching to find a good optimum for matching 2 large graphs. 
However, thanks to the local explainer, which produces concise subgraphs for important message-passing, the matching that is close to the best alignment can be efficiently computed.

\jure{CUT:}
However, if the input distribution for a class is multi-modal, or is a mixture of many distributions, there does not exist one computation graph prototype, but many prototypes corresponding to the class. Domain-specific techniques are required in such scenario, so that the alignment is done for each cluster of computation subgraphs corresponding to the label class.
\rex{do we even need to mention this to preempt reviewer questions?}

\xhdr{Prototype by alignment}
Once all adjacency matrices of nodes in class $c$ are aligned to the ordering defined by the reference adjacency matrix, we can then take the average of each edge to generate the prototype
$A_{\mathrm{proto}} = \frac{1}{n} \sum_i A_i$, where the $A_i$ is the aligned adjacency matrix for the important computation subgraph of a node of class $c$.

The prototype allows users to gain insights about common substructures identified for different nodes in the same class, and interpret the model prediction of a node by comparing its important computation subgraph to the class prototype.

\subsection{Extension to link prediction and graph classification tasks}
The framework of \name is general and can be adapted to explain the task of link prediction and graph classification as well. 
The same objective function in Equation \ref{eq:opt_general} can be modified in both scenario. 

In the task of link prediction between node $v_i$ and $v_j$, $Y$ is now the distribution of existence of links. The model prediction is:
\begin{equation}
    P_\Phi(Y | G_1 = A_{S_i} \odot M_1, G_2 = A_{S_j} \odot M2, X=X_S)
\end{equation}
\lexp now learns two mask to identify important computation subgraph for both $v_i$ and $v_j$ involved in the link prediction.

In the task of graph classification,
$Y$ is the distribution of label classes for graphs.
The model prediction is formulation is the same. However, $A_S$ in the objective is no longer the adjacency matrix for the computation graph of a particular node $v_i$, but the union of the computation graphs of all nodes in a graph $G$. In the case of GCN~\cite{kipf2016semi}, this is just the adjacency matrix of $G$.

In both scenarios. the optimization procedure remains the same.

\subsection{Training and efficiency}
The number of parameters in the optimization problem of the local model depends on the size of the adjacency matrix of the computation graph $A_c(v_i)$, since the mask parameter $M$ has the same size as $A_c(v_i)$. 
Most GCNs on large graphs use 2-3 hop neighborhoods~\CITE, the size of the computation graph is thus much smaller compared to the size of the entire graph, allowing the local model to be efficiently computed.

In very large graph systems such as PinSage~\cite{pinsage}, sampling or PageRank-based neighborhood is used. In these cases, the size of the computation graph is also relatively small, compared to the K-hop neighborhood.

In extreme cases such as relational reasoning~\cite{battaglia}, the computation graph for each node is the entire graph. However, typically in such model attention mechanism is employed~\cite{velickovic2018graph}. Based on the attention weights on edges, low-attention edges can be pruned to reduce the size of computation graph, thus allowing \lexp to be efficient.}
\section{Experiments}
\label{sec:exp}



\hide{
Results in Table 1:

0.925/0.815 = 13.5
0.925/0.882 = 4.9

0.836/0.739 = 13.1
0.836/0.750 = 11.4

0.948/0.824 = 15.0
0.948/0.905 = 4.7

0.875/0.612 = 43.0
0.875/0.667 = 31.2

Average = 17.1
}

We begin by describing the graphs, alternative baseline approaches, and experimental setup. We then present experiments on explaining \gnns for node classification and graph classification tasks. Our qualitative and quantitative analysis demonstrates that \name is accurate and effective in identifying explanations, both in terms of graph structure and node features. 

\xhdr{Synthetic datasets} 
We construct four kinds of node classification datasets (Table~1). (1) In \textsc{BA-Shapes}, we start with a base Barab\'{a}si-Albert (BA) graph on 300 nodes and a set of 80 five-node ``house''-structured network motifs, which are attached to randomly selected nodes of the base graph. The resulting graph is further perturbed by adding $0.1N$ random edges. 
Nodes are assigned to 4 classes based on their structural roles. 
In a house-structured motif, there are 3 types of roles: the top, middle and bottom node of the house. Therefore there are 4 different classes, corresponding to nodes at the top, middle, bottom of houses, and nodes that do not belong to a house.
(2) \textsc{BA-Community} dataset is a union of two \textsc{BA-Shapes} graphs. 
Nodes have normally distributed feature vectors and are assigned to one of 8 classes based on their structural roles and community memberships.
(3) In \textsc{Tree-Cycles}, we start with a base 8-level balanced binary tree and 80 six-node cycle motifs, which are attached to random nodes of the base graph.
(4) \textsc{Tree-Grid} is the same as \textsc{Tree-Cycles} except that 3-by-3 grid motifs are attached to the base tree graph in place of cycle motifs.

\xhdr{Real-world datasets}
We consider two graph classification datasets: (1) \textsc{Mutag} is a dataset of $4{,}337$ molecule graphs labeled according to their mutagenic effect on the Gram-negative bacterium \textit{S. typhimurium}~\citep{mutag}. 
(2)  \textsc{Reddit-Binary} is a dataset of $2{,}000$ graphs, each representing an online discussion thread on Reddit. In each graph, nodes are users participating in a thread, and edges indicate that one user replied to another user's comment. Graphs are labeled according to the type of user interactions in the thread: \textit{r/IAmA} and \textit{r/AskReddit} contain Question-Answer interactions, while 
\textit{r/TrollXChromosomes} and \textit{r/atheism} contain Online-Discussion interactions~\citep{yanardag2015deep}.

\begin{table*}[t]
    \centering
    \vspace{-4mm}
    \includegraphics[width=\linewidth]{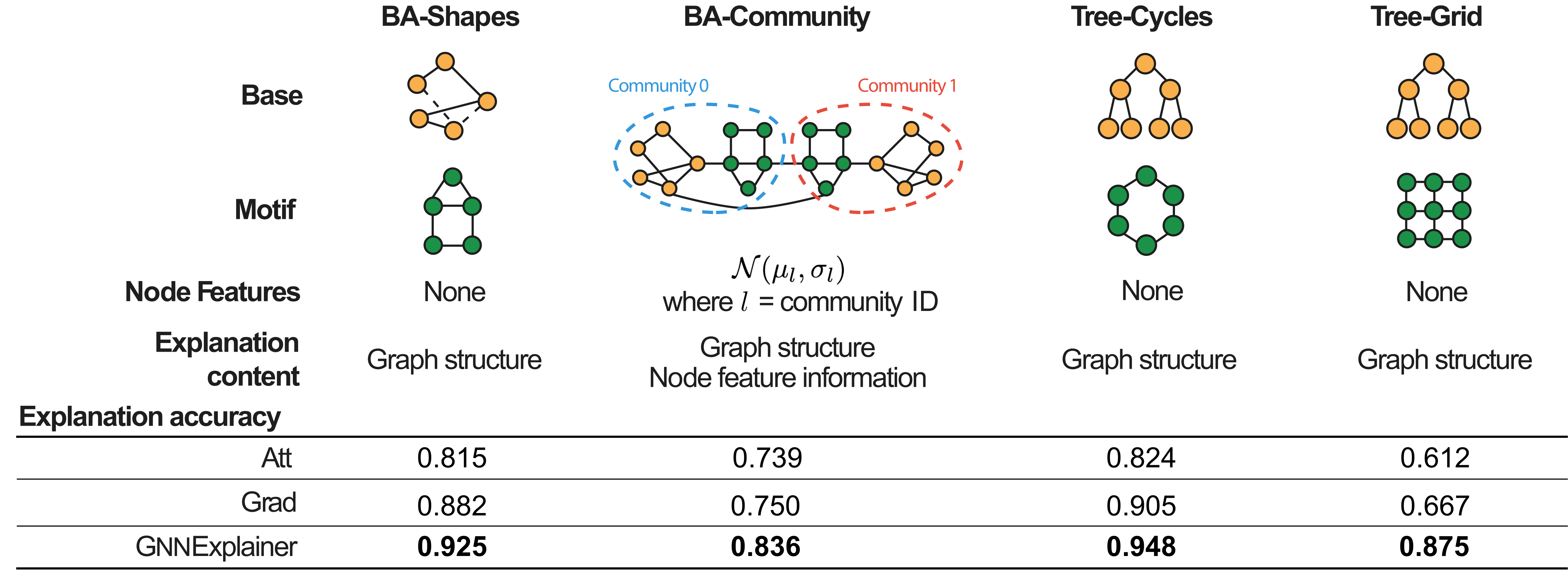}
    \caption{Illustration of synthetic datasets (refer to ``Synthetic datasets'' for details) together with performance evaluation of \name and alternative baseline explainability approaches.}
    \label{fig:synth_datasets}
    \vspace{-4mm}
\end{table*}

\hide{\marinka{(1) Consider adding vertical lines spanning the bottom three lines to make it look more like a table. (2) Can we get rid of node colors? It is confusing because colors do not represent node labels/classes. (3) Can we drop ``Number of classes''? It takes space. Also, it is not crucial for explainability? (4) Can we say ``No features'' instead ``Constant features''?}}

\hide{
We first construct four synthetic datasets to assess the approach for different aspects detailed in the desiderata in a controlled environment.These are detailed in Fig.~\ref{fig:synth_datasets}.
}

\cut{
\begin{enumerate}[leftmargin=*]
\item \xhdr{\textsc{BA-Shapes}} Given a base \ba (BA) graph of size $300$, a set of $80$ five-node house-structure network motifs are attached to random nodes. The resulting graph is further perturbed by uniformly randomly adding $0.1 n$ edges, where $n$ is the size of the graph. In order to isolate structural information gathered by the \gnn~ model, nodes are equipped with a set of constant features. Each node is assigned a label based on its role in the motif. There are $4$ label classes.
\item \xhdr{\textsc{BA-Community}} A union of $2$ \textsc{BA-Shapes} graphs. These two graphs are then randomly connected, forming two bridged communities. Nodes have normally distributed feature vectors. 
To test feature explanation together with structure explanation,
for dimension $2$, 
The mean of the feature distribution for different communities differs by $1$ standard deviation. 
The rest of the unimportant feature dimensions are drawn from the same distribution for all nodes in both communities.
The label of each node is based on both its structural role and its community; hence there are $8$ label classes.
\item \xhdr{\textsc{Tree-Cycles}} We construct a balanced binary tree of height $8$. 
A set of $80$ six-node cycle motifs are attached the same way as \textsc{BA-Shapes}.
This tests the model's ability to reach for multi-hop structural information in the assessment of a prediction.
This task is more challenging since the degree distributions for nodes in the tree and the motif are similar.
\item \xhdr{\textsc{Tree-Grid}} Same as \textsc{Tree-Cycles} except that a more complex $3$-by-$3$ grid is attached to the tree instead of cycles.
\end{enumerate}
}

\xhdr{Alternative baseline approaches} Many explainability methods cannot be directly applied to graphs (Section~\ref{sec:related}). Nevertheless, we here consider the following alternative approaches that can provide insights into predictions made by \gnns: (1) \textsc{Grad} is a gradient-based method. We compute gradient of the \gnn's loss function with respect to the adjacency matrix and the associated node features, similar to a saliency map approach. 
(2) \textsc{Att} is a graph attention \gnn (GAT)~\citep{velickovic2018graph} that learns attention weights for edges in the computation graph, which we use as a proxy measure of edge importance. While \textsc{Att} does consider graph structure, it does not explain using node features and can only explain GAT models.  
Furthermore, in \textsc{Att} it is not obvious which attention weights need to be used for edge importance, since a 1-hop neighbor of a node can also be a 2-hop neighbor of the same node due to cycles. Each edge's importance is thus computed as the average attention weight across all layers.

\xhdr{Setup and implementation details}
For each dataset, we first train a single \gnn for each dataset, and use \textsc{Grad} and \name to explain the predictions made by the \gnn. Note that the \textsc{Att} baseline requires using a graph attention architecture like GAT~\citep{velickovic2018graph}. We thus train a separate GAT model on the same dataset and use the learned edge attention weights for explanation.
Hyperparameters $K_M, K_F$ control the size of subgraph and feature explanations respectively, which is informed by prior knowledge about the dataset. For synthetic datasets, we set $K_M$ to be the size of ground truth.
On real-world datasets, we set $K_M = 10$.
We set $K_F = 5$ for all datasets.
We further fix our weight regularization hyperparameters across all node and graph classification experiments. We refer readers to the Appendix for more training details (Code and datasets are available at https://github.com/RexYing/gnn-model-explainer).

\hide{
To extract the explanation subgraph $G_S$, we first compute the importance weights on edges (gradients for \textsc{Grad} baseline, attention weights for \textsc{Att} baseline, and masked adjacency for \namelong). 
A threshold is used to remove low-weight edges, and identify the explanation subgraph $G_S$.
The ground truth explanations of all datasets are connected subgraphs. Therefore, we identify the explanation as the connected component containing the explained node in $G_S$. For graph classification, we identify the explanation by the maximum connected component of $G_S$.
For all methods, we perform a search to find the maximum threshold such that the explanation is at least of size $K$. When multiple edges have tied importance weights, all of them are included in the explanation.
}

\hide{
\begin{enumerate}[leftmargin=*]
\item \xhdr{\textsc{Grad}} Gradient-based method. We compute the gradient of model loss with respect to adjacency matrix and node features to be classified, and pick edges that have the highest absolute gradients, similar to saliency map. In graph classification, the gradient with respect to node features are averaged across nodes.
This method allows explaining important subgraphs as well as features.
\item \xhdr{\textsc{Att}} Graph Attention Network (GAT) provides explanation by attention weights on edges, serving as a proxy for edge importance in the computation graph~\cite{velickovic2018graph}. However, this method cannot be directly estimate feature importance, and cannot be used as post-hoc analysis for other \gnn models.
\end{enumerate}
}


\xhdr{Results}
We investigate questions: Does \name provide sensible explanations? How do explanations compare to the ground-truth knowledge? How does \name perform on various graph-based prediction tasks? Can it explain predictions made by different GNNs?

\begin{figure*}
    \centering
    \includegraphics[width=\textwidth]{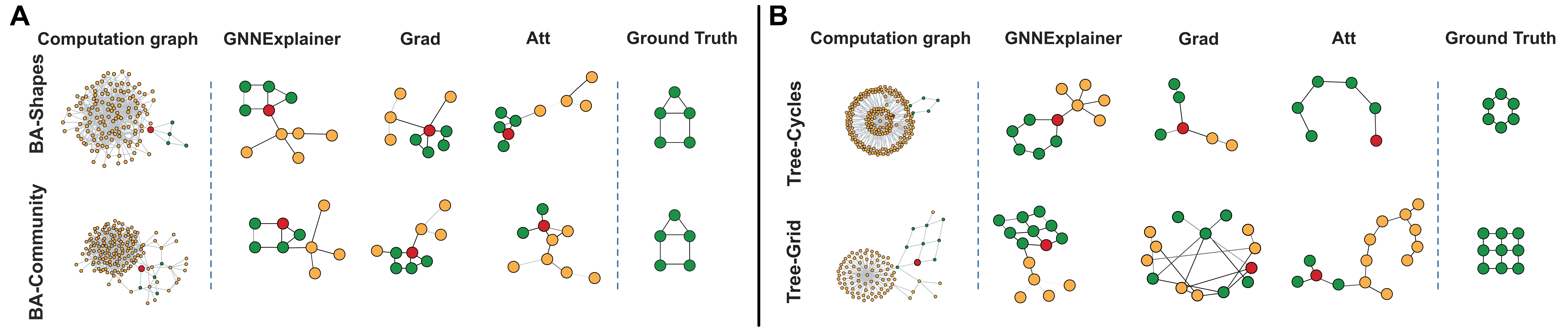}
    \vspace{-4mm}
    \caption{Evaluation of single-instance explanations. \textbf{A-B.} Shown are exemplar explanation subgraphs for node classification task on four synthetic datasets. Each method provides explanation for the red node's prediction.}
    \label{fig:subgraph_node}
\end{figure*}

\begin{figure*}
    \centering
    \includegraphics[width=\textwidth]{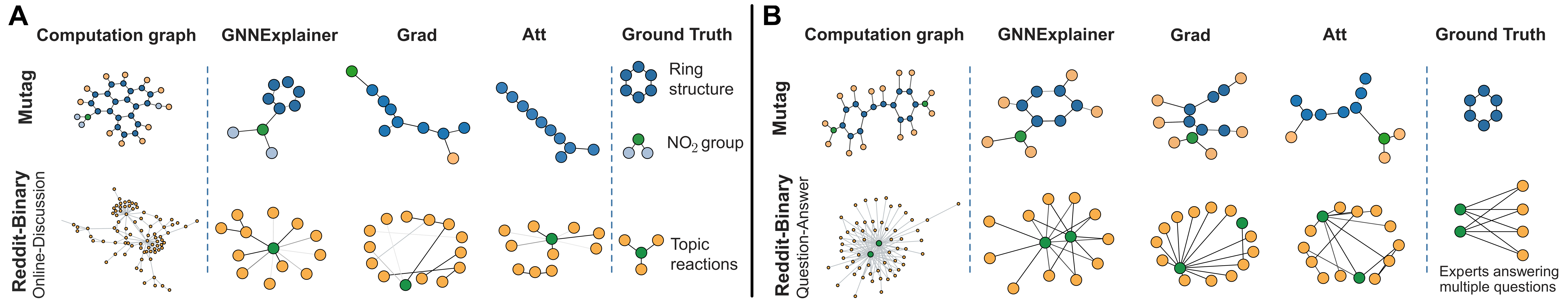}
    \vspace{-4mm}
    \caption{Evaluation of single-instance explanations. \textbf{A-B.} Shown are exemplar explanation subgraphs for graph classification task on two datasets, \textsc{Mutag} and \textsc{Reddit-Binary}.}
    \label{fig:subgraph_graph}
\end{figure*}

\xhdr{1) Quantitative analyses} 
Results on node classification datasets are shown in Table~\ref{fig:synth_datasets}. We have ground-truth explanations for synthetic datasets and we use them to calculate explanation accuracy for all explanation methods. Specifically, we formalize the explanation problem as a binary classification task, where edges in the ground-truth explanation are treated as labels and importance weights given by explainability method are viewed as prediction scores. A better explainability method predicts high scores for edges that are in the ground-truth explanation, and thus achieves higher explanation accuracy. 
Results show that \name outperforms alternative approaches by 17.1\% on average. Further, \name achieves up to 43.0\% higher accuracy on the hardest \textsc{Tree-Grid} dataset.

\begin{figure}
  \begin{minipage}[l]{0.5\textwidth}
    \includegraphics[width=\textwidth]{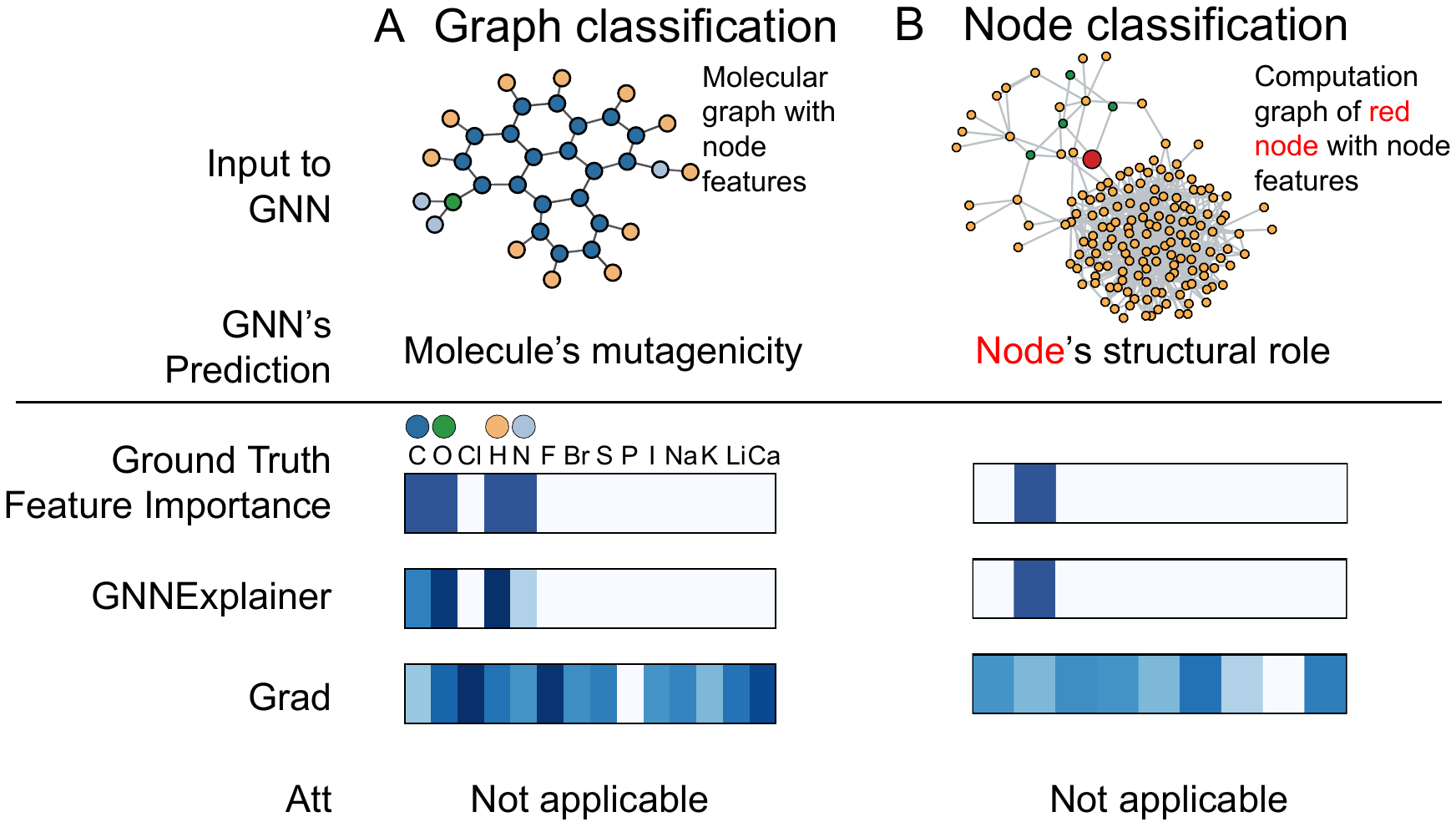}
  \end{minipage}\hfill
  \begin{minipage}[l]{0.48\textwidth}
    \caption{Visualization of features that are important for a GNN's prediction. \textbf{A.} Shown is a representative molecular graph from \textsc{Mutag} dataset (top). Importance of the associated graph features is visualized with a heatmap (bottom). In contrast with baselines, \name correctly identifies features that are important for predicting the molecule's mutagenicity, \ie C, O, H, and N atoms. \textbf{B.} Shown is a computation graph of a red node from \textsc{BA-Community} dataset (top). Again, \name successfully identifies the node feature that is important for predicting the structural role of the node but baseline methods fail.  \label{fig:feat_importance}}
  \end{minipage}
\end{figure}

\hide{
We also quantitatively evaluate \name on synthetic datasets where ground-truth explanations are available. Specifically, we formalize the explanation problem as a binary classification task, where edges in ground-truth explanations are treated as labels, while the importance weights given by different models are viewed as prediction scores. A good explanation model will predict high scores for edges in ground-truth explanations, thus will have better performance in this classification task. We compute the standard accuracy score for each model. As is shown in Table \ref{fig:synth_datasets}, \name significantly out-performs the baselines by an average of 9.5\%.

The explanation should also highlight relevant feature information, not only from itself but even within the set of neighbours that propagated on its most influential message-passing pathways\footnote{Feature explanations are shown for the two datasets with node features, \ie, \textsc{Mutag} and \textsc{BA-Community}.}. In an experiment such as \textsc{BA-Comm}, the explainer is forced to also integrate information from a restricted number of feature dimensions. While \namelong indeed highlights a compact feature representation in Figure~\ref{fig:feat_importance}, gradient-based approaches struggle to cope with the added noise, giving high importance scores to irrelevant feature dimensions. We also re-iterate the model's ability to learn structure and feature-based information jointly in Figure~\ref{fig:subgraph}, where \namelong is again able to identify the correct substructure, as is the Attention-based baseline.
}

\xhdr{2) Qualitative analyses} 
%
Results are shown in Figures~\ref{fig:subgraph_node}--\ref{fig:feat_importance}. In a topology-based prediction task with no node features, \emph{e.g.} \textsc{BA-Shapes} and \textsc{Tree-Cycles}, \namelong correctly identifies network motifs that explain node labels, \ie  structural labels (Figure~\ref{fig:subgraph_node}). 
As illustrated in the figures, house, cycle and tree motifs are identified by \name but not by baseline methods. 
In Figure~\ref{fig:subgraph_graph}, we investigate explanations for graph classification task.
In \textsc{Mutag} example, colors indicate node features, which represent atoms (hydrogen H, carbon C, \textit{etc}). \name correctly identifies carbon ring as well as chemical groups $NH_2$ and $NO_2$, which are known to be mutagenic~\cite{mutag}. 

Further, in \textsc{Reddit-Binary} example, we see that Question-Answer graphs (2nd row in Figure~\ref{fig:subgraph_graph}B) have 2-3 high degree nodes that simultaneously connect to many low degree nodes, which makes sense because in QA threads on Reddit we typically have 2-3 experts who all answer many different questions~\citep{kumar2018community}.
Conversely, we observe that discussion patterns commonly exhibit tree-like patterns (2nd row in Figure~\ref{fig:subgraph_graph}A), since a thread on Reddit is usually a reaction to a single topic~\citep{kumar2018community}. 
On the other hand, \textsc{Grad} and \textsc{Att} methods give incorrect or incomplete explanations. For example, both baseline methods miss cycle motifs in \textsc{Mutag} dataset and more complex grid motifs in \textsc{Tree-Grid} dataset.
Furthermore, although edge attention weights in \textsc{Att} can be interpreted as importance scores for message passing, the weights are shared across all nodes in input the graph, and as such \textsc{Att} fails to provide high quality single-instance explanations.

An essential criterion for explanations is that they must be interpretable, \ie, provide a qualitative understanding of the relationship between the input nodes and the prediction. Such a requirement implies that explanations should be easy to understand while remaining exhaustive. This means that a \gnn explainer should take into account both the structure of the underlying graph as well as the associated features when they are available. 
Figure~\ref{fig:feat_importance} shows results of an experiment in which \name jointly considers structural information as well as information from a small number of feature dimensions\footnote{Feature explanations are shown for the two datasets with node features, \ie, \textsc{Mutag} and \textsc{BA-Community}.}. While \namelong indeed highlights a compact feature representation in Figure~\ref{fig:feat_importance}, gradient-based approaches struggle to cope with the added noise, giving high importance scores to irrelevant feature dimensions. 


Further experiments on multi-instance explanations using graph prototypes are in Appendix.

\hide{
First, we show the explainer's local edge fidelity (single-instance explanations) through its ability to highlight relevant message-passing pathways. 
In a strictly topology-based prediction task such \textsc{BA-Shapes} and \textsc{Tree-Cycles}, \namelong correctly identifies the motifs required for accurate classification (top 2 rows of Figure~\ref{fig:subgraph_node}). 
As illustrated, \namelong looks for a concise subgraph of the computation graph that best explains the prediction for a single queried node. 
All of house, cycle and tree motifs are correctly identified in almost all instances. 

In Figure~\ref{fig:subgraph_node}, we compare explanations generated for graph classification models.
For the \textsc{Mutag} example, colors indicate node features, which represent the atom in question (hydrogen, carbon, \textit{etc}). Here, the carbon ring, as well as the functional groups $NH_2$ and $NO_2$, that are known to be mutagenic, are correctly identified~\cite{mutag}. 
In the \textsc{Reddit-Binary} dataset, QA graphs ($2$-nd row to the right) have 2-3 high degree nodes that simultaneously connect to many low degree nodes, due to 2-3 experts all answering many different questions.
In comparison, we observe that discussion patterns commonly exhibit tree-like patterns, since the threads are usually reactions to a single topic. The second row of Figure~\ref{fig:subgraph_node} shows that such pattern is consistently identified by \namelong.

On the other hand, the Gradient- and Attention-based methods often give incorrect or incomplete patterns as an explanation. For example, they miss the cycles in \textsc{Mutag} or the more complex grid structure of \textsc{Tree-Grid}.
Furthermore, although an edge's attention weights in the \textsc{Att} baseline can be interpreted as its importance in message passing, the graph attention weights are shared for all nodes in the graph. Therefore it fails to provide high quality single instance explanations.


\marinka{An essential criterion for explanations is that they must be interpretable, \ie, provide a qualitative understanding of the relationship between the input nodes and the prediction. Such a requirement implies that explanations should be easy to understand while remaining exhaustive. This means that a \gnn explainer should take into account both the structure of the underlying graph $G$ as well as the associated features, if they are available. ...}

Further experiments on multi-instance explanations using graph prototypes are in the Appendix.
}

\hide{
\subsection{Multi-instance prototype}

In the context of multi-instance explanations, an explainer must not only highlight information locally relevant to a particular prediction, but also help emphasize higher-level correlations across instances. These instances can be related in arbitrary ways, but the most evident is class-membership. The assumption is that members of a class share common characteristics, and the model should help highlight them. For example, mutagenic compounds are often found to have certain characteristic functional groups that such $NO_2$, a pair of Oxygen atoms with a Nitrogen. While Figure~\ref{fig:subgraph} already hints to their presence, which the trained eye might recognize. The evidence grows stronger when a prototype is generated by \namelong, shown in Figure~\ref{fig:prototype}. The model is able to pick-up on this functional structure, and promote it as archetypal of mutagenic compounds.

\begin{figure}
    \centering
    \includegraphics[width=0.8\columnwidth]{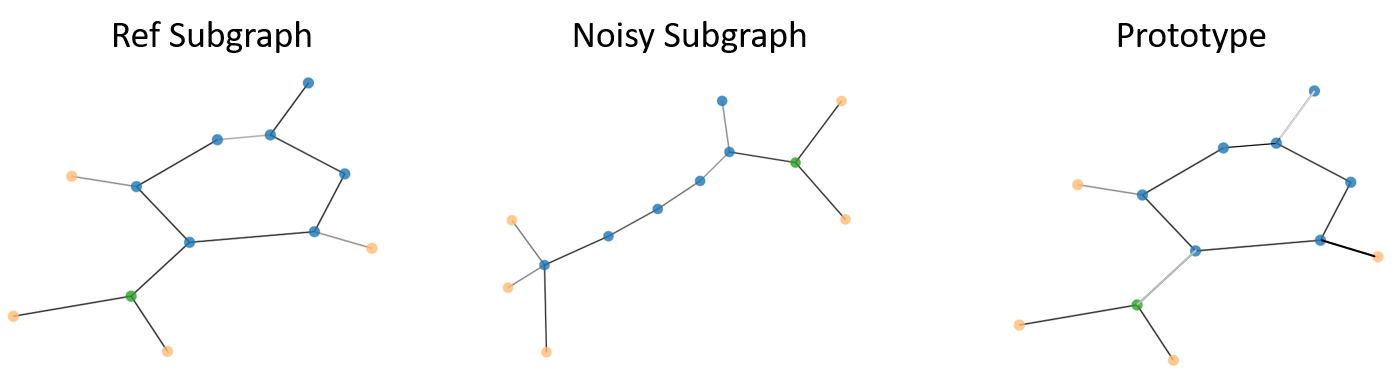}
    \caption{\longname is able to provide a prototype for a given node class, which can help identify functional subgraphs, e.g. a mutagenic compound from the \textsc{Mutag} dataset.}
    \label{fig:prototype}
    \vspace{-3mm}
\end{figure}
}



\cut{
\begin{table}[t]
\centering
\begin{footnotesize}
\caption{\namelong compared to baselines in identifying subgraphs using AUC. \marinka{Add a column with Tree-Grid results? Mutag as well?}}
\label{tab:results_pr}
\begin{tabular}{@{}llll@{}}
\toprule
              & \textsc{BA-Shapes} & \textsc{BA-Community} & \textsc{Tree-Cycles} & \textsc{Tree-Grid} \\ 
GAT      & $81.5$ & $0.73.9$  & $ 82.4$   &   $61.2$   \\
Grad       & $88.2$ & $0.75.0$  & $ 90.5$ & $66,7$ \\
\gnn  & $\mathbf{92.5}$ & $\mathbf{0.83.6}$  & $\mathbf{0.94.8}$  & $\mathbf{76.1}$ \\
\end{tabular}
\end{footnotesize}
    \vspace{-6mm}

\end{table}

\begin{table}[t]
\centering
\begin{footnotesize}
\caption{\namelong compared to \textsc{Grad} baseline in feature importance map using cross entropy.}
\label{tab:results_pr}
\begin{tabular}{@{}llll@{}}
\toprule
              & \textsc{Mutag} & \textsc{BA-Community} & \textsc{Tree-Cycles} \\ 
Grad       & $0.907$ & $0.183$  & $0.785$ \\
\gnn     & $\mathbf{0.221}$ & $\mathbf{0.003}$  & $\mathbf{0.024}$   \\
\end{tabular}
\end{footnotesize}
    \vspace{-4mm}

\end{table}
}

\cut{
\begin{figure*}
    \centering
    \begin{subfigure}[b]{\columnwidth}
    \includegraphics[width=\textwidth]{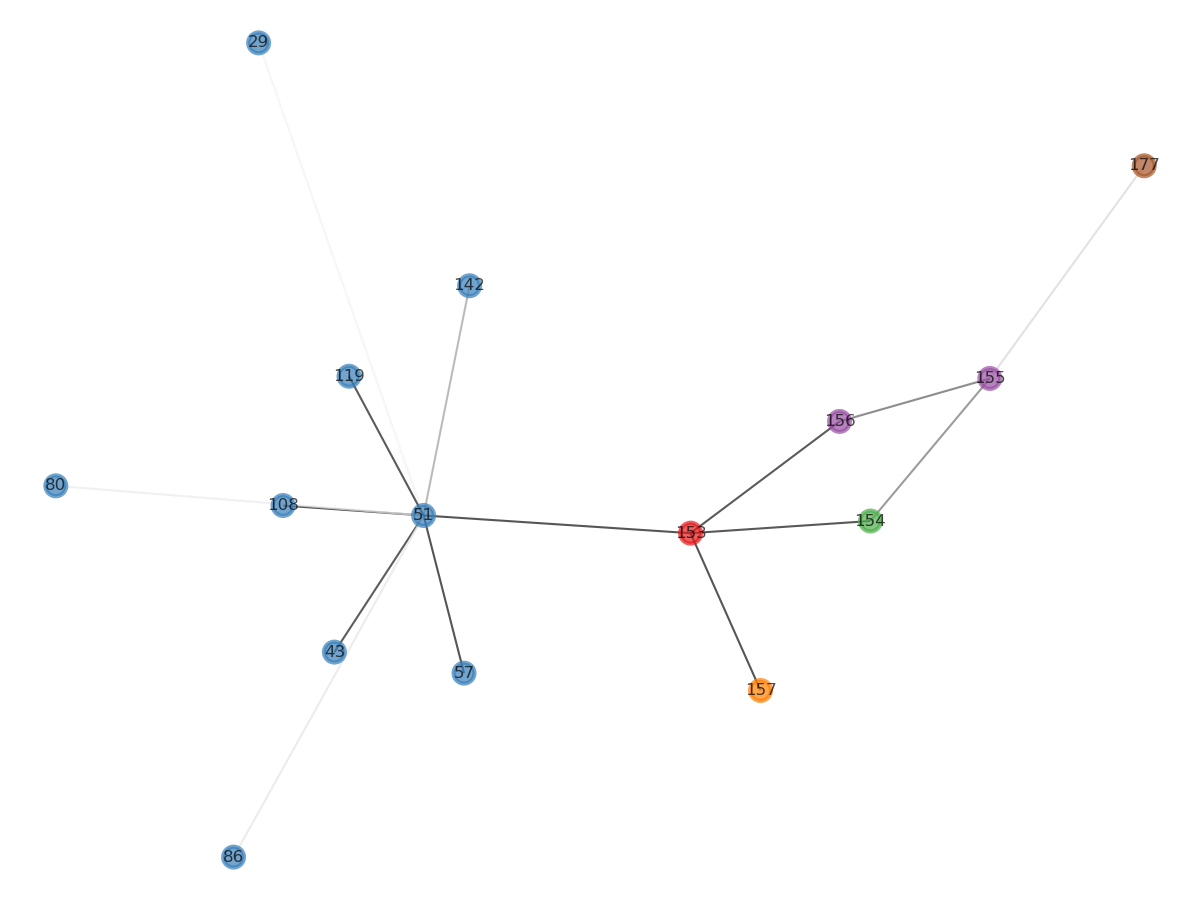}
    \end{subfigure}
    \begin{subfigure}[b]{\columnwidth}
        \includegraphics[width=\textwidth]{figs/grad_subgraph}
    \end{subfigure}
    \caption{Caption}
    \label{fig:my_label}
\end{figure*}
}
\section{Conclusion}
\label{sec:conclusion}

We present \longname, a novel method for explaining predictions of any GNN on any graph-based machine learning task without requiring modification of the underlying GNN architecture or re-training. 
We show how \name can leverage recursive neighborhood-aggregation scheme of graph neural networks to identify important graph pathways as well as highlight relevant node feature information that is passed along edges of the pathways.
While the problem of explainability of machine-learning predictions has received substantial attention in recent literature, our work is unique in the sense that it presents an approach that operates on relational structures---graphs with rich node features---and provides a straightforward interface for making sense out of GNN predictions, debugging GNN models, and identifying systematic patterns of mistakes.

\hide{
In this paper, we present \longname, a novel approach for explaining predictions of any GNN on any graph-based machine learning task. 
which provides insights into any \gnn that satisfies the neural message-passing scheme, without any modification to its architecture or re-training, and across any prediction task on graphs. 
It is able to leverage the recursive neighborhood-aggregation scheme of graph neural networks to identify not only important computational pathways but also highlight the relevant feature information that is passed along these edges. 
}

\subsubsection*{Acknowledgments}
Jure Leskovec is a Chan Zuckerberg Biohub investigator.
We gratefully acknowledge the support of 
DARPA under FA865018C7880 (ASED) and MSC; 
NIH under No. U54EB020405 (Mobilize); 
ARO under No. 38796-Z8424103 (MURI); 
IARPA under No. 2017-17071900005 (HFC),
NSF under No. OAC-1835598 (CINES) and HDR; 
Stanford Data Science Initiative, 
Chan Zuckerberg Biohub, 
JD.com, Amazon, Boeing, Docomo, Huawei, Hitachi, Observe, Siemens, UST Global.
The U.S. Government is authorized to reproduce and distribute reprints for Governmental purposes notwithstanding any copyright notation thereon. Any opinions, findings, and conclusions or recommendations expressed in this material are those of the authors and do not necessarily reflect the views, policies, or endorsements, either expressed or implied, of DARPA, NIH, ONR, or the U.S. Government.



%
\bibliographystyle{plain}
\bibliography{refs}

\begin{thebibliography}{10}

\bibitem{adadi_peeking_2018}
A.~Adadi and M.~Berrada.
\newblock Peeking {Inside} the {Black}-{Box}: {A} {Survey} on {Explainable}
  {Artificial} {Intelligence} ({XAI}).
\newblock {\em IEEE Access}, 6:52138--52160, 2018.

\bibitem{2018sanity}
J.~Adebayo, J.~Gilmer, M.~Muelly, I.~Goodfellow, M.~Hardt, and B.~Kim.
\newblock Sanity checks for saliency maps.
\newblock In {\em NeurIPS}, 2018.

\bibitem{augasta_reverse_2012}
M.~Gethsiyal Augasta and T.~Kathirvalavakumar.
\newblock Reverse {Engineering} the {Neural} {Networks} for {Rule} {Extraction}
  in {Classification} {Problems}.
\newblock {\em Neural Processing Letters}, 35(2):131--150, April 2012.

\bibitem{battaglia}
Peter~W Battaglia, Jessica~B Hamrick, Victor Bapst, Alvaro Sanchez-Gonzalez,
  Vinicius Zambaldi, Mateusz Malinowski, Andrea Tacchetti, David Raposo, Adam
  Santoro, Ryan Faulkner, et~al.
\newblock Relational inductive biases, deep learning, and graph networks.
\newblock {\em arXiv:1806.01261}, 2018.

\bibitem{Chen2018StochasticTO}
J.~Chen, J.~Zhu, and L.~Song.
\newblock Stochastic training of graph convolutional networks with variance
  reduction.
\newblock In {\em ICML}, 2018.

\bibitem{chen2018learning}
Jianbo Chen, Le~Song, Martin~J Wainwright, and Michael~I Jordan.
\newblock Learning to explain: An information-theoretic perspective on model
  interpretation.
\newblock {\em arXiv preprint arXiv:1802.07814}, 2018.

\bibitem{fastgcn}
Jie Chen, Tengfei Ma, and Cao Xiao.
\newblock Fastgcn: fast learning with graph convolutional networks via
  importance sampling.
\newblock In {\em ICLR}, 2018.

\bibitem{chen2018supervised}
Z.~Chen, L.~Li, and J.~Bruna.
\newblock Supervised community detection with line graph neural networks.
\newblock In {\em ICLR}, 2019.

\bibitem{cho2011friendship}
E.~Cho, S.~Myers, and J.~Leskovec.
\newblock Friendship and mobility: user movement in location-based social
  networks.
\newblock In {\em KDD}, 2011.

\bibitem{mutag}
A.~Debnath et~al.
\newblock Structure-activity relationship of mutagenic aromatic and
  heteroaromatic nitro compounds. correlation with molecular orbital energies
  and hydrophobicity.
\newblock {\em Journal of Medicinal Chemistry}, 34(2):786--797, 1991.

\bibitem{doshi-velez_towards_2017}
F.~Doshi-Velez and B.~Kim.
\newblock Towards {A} {Rigorous} {Science} of {Interpretable} {Machine}
  {Learning}.
\newblock 2017.
\newblock arXiv: 1702.08608.

\bibitem{duvenaud_convolutional_2015}
D.~Duvenaud et~al.
\newblock Convolutional networks on graphs for learning molecular fingerprints.
\newblock In {\em NIPS}, 2015.

\bibitem{Erhan2009VisualizingHF}
D.~Erhan, Y.~Bengio, A.~Courville, and P.~Vincent.
\newblock Visualizing higher-layer features of a deep network.
\newblock {\em University of Montreal}, 1341(3):1, 2009.

\bibitem{fisher_all_2018}
A.~Fisher, C.~Rudin, and F.~Dominici.
\newblock All {Models} are {Wrong} but many are {Useful}: {Variable}
  {Importance} for {Black}-{Box}, {Proprietary}, or {Misspecified} {Prediction}
  {Models}, using {Model} {Class} {Reliance}.
\newblock January 2018.
\newblock arXiv: 1801.01489.

\bibitem{guidotti_survey_2018}
R.~Guidotti et~al.
\newblock A {Survey} of {Methods} for {Explaining} {Black} {Box} {Models}.
\newblock {\em ACM Comput. Surv.}, 51(5):93:1--93:42, 2018.

\bibitem{graphsage}
W.~Hamilton, Z.~Ying, and J.~Leskovec.
\newblock Inductive representation learning on large graphs.
\newblock In {\em NIPS}, 2017.

\bibitem{hooker_discovering_2004}
G.~Hooker.
\newblock Discovering additive structure in black box functions.
\newblock In {\em {KDD}}, 2004.

\bibitem{Huang2018AdaptiveST}
W.B. Huang, T.~Zhang, Y.~Rong, and J.~Huang.
\newblock Adaptive sampling towards fast graph representation learning.
\newblock In {\em NeurIPS}, 2018.

\bibitem{Kang2019explaine}
Bo~Kang, Jefrey Lijffijt, and Tijl De~Bie.
\newblock Explaine: An approach for explaining network embedding-based link
  predictions.
\newblock {\em arXiv:1904.12694}, 2019.

\bibitem{kingma2013auto}
Diederik~P Kingma and Max Welling.
\newblock Auto-encoding variational bayes.
\newblock In {\em NeurIPS}, 2013.

\bibitem{kipf2016semi}
T.~N. Kipf and M.~Welling.
\newblock Semi-supervised classification with graph convolutional networks.
\newblock In {\em ICLR}, 2016.

\bibitem{kipf2018neural}
Thomas Kipf, Ethan Fetaya, Kuan-Chieh Wang, Max Welling, and Richard Zemel.
\newblock Neural relational inference for interacting systems.
\newblock In {\em ICML}, 2018.

\bibitem{koh_understanding_2017}
P.~W. Koh and P.~Liang.
\newblock Understanding black-box predictions via influence functions.
\newblock In {\em ICML}, 2017.

\bibitem{kumar2018community}
Srijan Kumar, William~L Hamilton, Jure Leskovec, and Dan Jurafsky.
\newblock Community interaction and conflict on the web.
\newblock In {\em WWW}, pages 933--943, 2018.

\bibitem{lakkaraju_interpretable_2017}
H.~Lakkaraju, E.~Kamar, R.~Caruana, and J.~Leskovec.
\newblock Interpretable \& {Explorable} {Approximations} of {Black} {Box}
  {Models}, 2017.

\bibitem{li2015gated}
Y.~Li, D.~Tarlow, M.~Brockschmidt, and R.~Zemel.
\newblock Gated graph sequence neural networks.
\newblock {\em arXiv:1511.05493}, 2015.

\bibitem{lundberg_unified_2017}
S.~Lundberg and Su-In Lee.
\newblock A {Unified} {Approach} to {Interpreting} {Model} {Predictions}.
\newblock In {\em NIPS}, 2017.

\bibitem{neil2018interpretable}
D.~Neil et~al.
\newblock Interpretable {Graph} {Convolutional} {Neural} {Networks} for
  {Inference} on {Noisy} {Knowledge} {Graphs}.
\newblock In {\em ML4H Workshop at NeurIPS}, 2018.

\bibitem{ribeiro_why_2016}
M.~Ribeiro, S.~Singh, and C.~Guestrin.
\newblock Why should i trust you?: Explaining the predictions of any
  classifier.
\newblock In {\em KDD}, 2016.

\bibitem{schmitz_ann-dt:_1999}
G.~J. Schmitz, C.~Aldrich, and F.~S. Gouws.
\newblock {ANN}-{DT}: an algorithm for extraction of decision trees from
  artificial neural networks.
\newblock {\em IEEE Transactions on Neural Networks}, 1999.

\bibitem{shrikumar_learning_2017}
A.~Shrikumar, P.~Greenside, and A.~Kundaje.
\newblock Learning {Important} {Features} {Through} {Propagating} {Activation}
  {Differences}.
\newblock In {\em ICML}, 2017.

\bibitem{sundararajan_axiomatic_nodate}
M.~Sundararajan, A.~Taly, and Q.~Yan.
\newblock Axiomatic {Attribution} for {Deep} {Networks}.
\newblock In {\em ICML}, 2017.

\bibitem{velickovic2018graph}
P.~Velickovic, G.~Cucurull, A.~Casanova, A.~Romero, P.~Li{\`o}, and Y.~Bengio.
\newblock Graph attention networks.
\newblock In {\em ICLR}, 2018.

\bibitem{PhysRevLett.120.145301}
T.~Xie and J.~Grossman.
\newblock Crystal graph convolutional neural networks for an accurate and
  interpretable prediction of material properties.
\newblock In {\em Phys. Rev. Lett.}, 2018.

\bibitem{xu2018powerful}
K.~Xu, W.~Hu, J.~Leskovec, and S.~Jegelka.
\newblock How powerful are graph neural networks?
\newblock In {\em ICRL}, 2019.

\bibitem{xujumping}
K.~Xu, C.~Li, Y.~Tian, T.~Sonobe, K.~Kawarabayashi, and S.~Jegelka.
\newblock Representation learning on graphs with jumping knowledge networks.
\newblock In {\em ICML}, 2018.

\bibitem{yanardag2015deep}
Pinar Yanardag and SVN Vishwanathan.
\newblock Deep graph kernels.
\newblock In {\em KDD}, pages 1365--1374. ACM, 2015.

\bibitem{DBLP:journals/corr/abs-1811-09720}
C.~Yeh, J.~Kim, I.~Yen, and P.~Ravikumar.
\newblock Representer point selection for explaining deep neural networks.
\newblock In {\em NeurIPS}, 2018.

\bibitem{pinsage}
R.~Ying, R.~He, K.~Chen, P.~Eksombatchai, W.~Hamilton, and J.~Leskovec.
\newblock Graph convolutional neural networks for web-scale recommender
  systems.
\newblock In {\em KDD}, 2018.

\bibitem{ying2018hierarchical}
Z.~Ying, J.~You, C.~Morris, X.~Ren, W.~Hamilton, and J.~Leskovec.
\newblock Hierarchical graph representation learning with differentiable
  pooling.
\newblock In {\em NeurIPS}, 2018.

\bibitem{you2018graph}
J.~You, B.~Liu, R.~Ying, V.~Pande, and J.~Leskovec.
\newblock Graph convolutional policy network for goal-directed molecular graph
  generation.
\newblock 2018.

\bibitem{You2019position}
J.~You, Rex Ying, and J.~Leskovec.
\newblock Position-aware graph neural networks.
\newblock In {\em ICML}, 2019.

\bibitem{fleet_visualizing_2014}
M.~Zeiler and R.~Fergus.
\newblock Visualizing and {Understanding} {Convolutional} {Networks}.
\newblock In {\em ECCV}. 2014.

\bibitem{zhang2018link}
M.~Zhang and Y.~Chen.
\newblock Link prediction based on graph neural networks.
\newblock In {\em NIPS}, 2018.

\bibitem{zhang_deep_2018}
Z.~Zhang, Peng C., and W.~Zhu.
\newblock Deep {Learning} on {Graphs}: {A} {Survey}.
\newblock {\em arXiv:1812.04202}, 2018.

\bibitem{zhou_graph_2018}
J.~Zhou, G.~Cui, Z.~Zhang, C.~Yang, Z.~Liu, and M.~Sun.
\newblock Graph {Neural} {Networks}: {A} {Review} of {Methods} and
  {Applications}.
\newblock {\em arXiv:1812.08434}, 2018.

\bibitem{calders_deepred_2016}
J.~Zilke, E.~Loza~Mencia, and F.~Janssen.
\newblock {DeepRED} - {Rule} {Extraction} from {Deep} {Neural} {Networks}.
\newblock In {\em Discovery {Science}}. Springer International Publishing,
  2016.

\bibitem{zintgraf2017visualizing}
L.~Zintgraf, T.~Cohen, T.~Adel, and M.~Welling.
\newblock Visualizing deep neural network decisions: Prediction difference
  analysis.
\newblock In {\em ICLR}, 2017.

\bibitem{zitnik2018decagon}
M.~Zitnik, M.~Agrawal, and J.~Leskovec.
\newblock Modeling polypharmacy side effects with graph convolutional networks.
\newblock {\em Bioinformatics}, 34, 2018.

\end{thebibliography}

%
\appendix
\section{Multi-instance explanations}

The problem of multi-instance explanations for graph neural networks is challenging and an important area to study.

Here we propose a solution based on \name to find common components of explanations for a set of 10 explanations for 10 different instances in the same label class. More research in this area is necessary to design efficient Multi-instance explanation methods.
The main challenges in practice is mainly due to the difficulty to perform graph alignment under noise and variances of node neighborhood structures for nodes in the same class. The problem is closely related to finding the maximum common subgraphs of explanation graphs, which is an NP-hard problem. In the following we introduces a neural approach to this problem. However, note that existing graph libraries (based on heuristics or integer programming relaxation) to find the maximal common subgraph of graphs can be employed to replace the neural components of the following procedure, when trying to identify and align with a prototype.

The output of a single-instance \name indicates what graph structural and node feature information  is important for a given prediction. To obtain an understanding of ``why is a given set of nodes classified with label $y$'', we want to also obtain a global explanation of the class, which can shed light on how the identified structure for a given node is related to a prototypical structure unique for its label. To this end, we propose an alignment-based multi-instance \name. 

For any given class, we first choose a reference node. Intuitively, this node should be a prototypical node for the class. Such node can be found by computing the mean of the embeddings of all nodes in the class, and choose the node whose embedding is the closest to the mean. Alternatively, if one has prior knowledge about the important computation subgraph, one can choose one which matches most to the prior knowledge.

Given the reference node for class $c$, $v_c$, and its associated important computation subgraph $G_S(v_c)$, 
we align each of the identified computation subgraphs for all nodes in class $c$ to the reference $G_S(v_c)$.
Utilizing the idea in the context of differentiable pooling~\cite{ying2018hierarchical}, we use the a relaxed alignment matrix to find correspondence between nodes in an computation subgraph $G_S(v)$ and nodes in the reference computation subgraph $G_S(v_c)$.
Let $A_v$ and $X_v$ be the adjacency matrix and the associated feature matrix of the to-be-aligned computation subgraph. Similarly let $A^*$ be the adjacency matrix and associated feature matrix of the reference computation subgraph.
Then we optimize the relaxed alignment matrix $P \in \mathbb{R}^{n_v \times n^*}$, where $n_v$ is the number of nodes in $G_S(v)$, and $n^*$ is the number of nodes in $G_S(v_c)$ as follows:
\begin{equation}
    \label{eq:glob}
    \min_P | P^T A_v P - A^*| + |P^T X_v - X^*|.
\end{equation}
The first term in Eq.~(\ref{eq:glob}) specifies that after alignment, the aligned adjacency for $G_S(v)$ should be as close to $A^*$ as possible. The second term in the equation specifies that the features should for the aligned nodes should also be close. 

In practice, it is often non-trivial for the relaxed graph matching to find a good optimum for matching 2 large graphs. 
However, thanks to the single-instance explainer, which produces concise subgraphs for important message-passing, a matching that is close to the best alignment can be efficiently computed.



\xhdr{Prototype by alignment}
We align the adjacency matrices of all nodes in class $c$, such that they are aligned with respect to the ordering defined by the reference adjacency matrix. We then use median to generate a prototype that is resistent to outliers,
$A_{\mathrm{proto}} = \mathrm{median}( A_i)$, where $A_i$ is the aligned adjacency matrix representing explanation for $i$-th node in class $c$.
Prototype $A_{\mathrm{proto}}$ allows users to gain insights into structural graph patterns shared between nodes that belong to the same class. Users can then investigate a particular node by comparing its explanation to the class prototype.

\section{Experiments on multi-instance explanations and prototypes}

In the context of multi-instance explanations, an explainer must not only highlight information locally relevant to a particular prediction, but also help emphasize higher-level correlations across instances. These instances can be related in arbitrary ways, but the most evident is class-membership. The assumption is that members of a class share common characteristics, and the model should help highlight them. For example, mutagenic compounds are often found to have certain characteristic functional groups that such $NO_2$, a pair of Oxygen atoms together with a Nitrogen atom. A trained eye might notice that Figure~\ref{fig:prototype} already hints at their presence. The evidence grows stronger when a prototype is generated by \namelong, shown in Figure~\ref{fig:prototype}. The model is able to pick-up on this functional structure, and promote it as archetypal of mutagenic compounds.

\begin{figure}
    \centering
    \includegraphics[width=0.8\columnwidth]{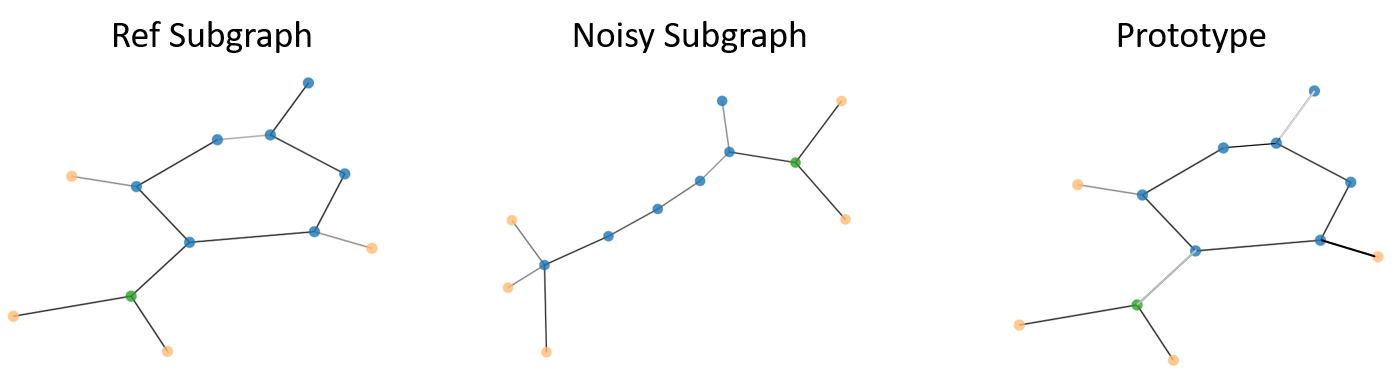}
    \caption{\longname is able to provide a prototype for a given node class, which can help identify functional subgraphs, e.g. a mutagenic compound from the \textsc{Mutag} dataset.}
    \label{fig:prototype}
    \vspace{-3mm}
\end{figure}

\section{Further implementation details}

\xhdr{Training details}
We use the Adam optimizer to train both the GNN and explaination methods. All GNN models are trained for 1000 epochs with learning rate 0.001, reaching accuracy of at least 85\% for graph classification datasets, and 95\% for node classification datasets. The train/validation/test split is $80/10/10\%$ for all datasets.
In \name, we use the same optimizer and learning rate, and train for 100 - 300 epochs. This is efficient since \name only needs to be trained on a local computation graph with $<100$ nodes.

\xhdr{Regularization}
In addition to graph size constraint and graph laplacian constraint, we further impose the feature size constraint, which constrains that the number of unmasked features do not exceed a threshold.
The regularization hyperparameters for subgraph size is $0.005$; for laplacian is $0.5$; for feature explanation is $0.1$.
The same values of hyperparameters are used across all experiments.

\xhdr{Subgraph extraction}
To extract the explanation subgraph $G_S$, we first compute the importance weights on edges (gradients for \textsc{Grad} baseline, attention weights for \textsc{Att} baseline, and masked adjacency for \namelong). 
A threshold is used to remove low-weight edges, and identify the explanation subgraph $G_S$.
The ground truth explanations of all datasets are connected subgraphs. Therefore, we identify the explanation as the connected component containing the explained node in $G_S$. For graph classification, we identify the explanation by the maximum connected component of $G_S$.
For all methods, we perform a search to find the maximum threshold such that the explanation is at least of size $K_M$. When multiple edges have tied importance weights, all of them are included in the explanation.

\end{document}